\documentclass[10pt,twocolumn,letterpaper]{article}

\usepackage[pagenumbers]{cvpr} 

\usepackage{graphicx}
\usepackage{amsmath}
\usepackage{amssymb}
\usepackage{booktabs}
\usepackage{multirow}
\usepackage{threeparttable}
\usepackage{caption}
\usepackage{bbding}
\usepackage{pifont}
\usepackage{color}
\usepackage{tabularx}
\usepackage{xcolor}
\usepackage{arydshln}
\newcommand{\cmark}{\ding{51}}%
\newcommand{\xmark}{\ding{55}}%

\usepackage[pagebackref,breaklinks,colorlinks]{hyperref}

\usepackage[capitalize]{cleveref}
\crefname{section}{Sec.}{Secs.}
\Crefname{section}{Section}{Sections}
\Crefname{table}{Table}{Tables}
\crefname{table}{Tab.}{Tabs.}

\def\cvprPaperID{5669} 
\def\confName{CVPR}
\def\confYear{2023}

\begin{document}

\title{SFD2: Semantic-guided Feature Detection and Description}
\author{Fei Xue \quad Ignas Budvytis \quad Roberto Cipolla \\
	{ University of Cambridge}  \\
	{\small \{fx221, ib255, rc10001\}@cam.ac.uk }
}

\maketitle


\begin{abstract}	
	Visual localization is a fundamental task for various applications including autonomous driving and robotics. Prior methods focus on extracting large amounts of often redundant locally reliable features, resulting in limited efficiency and accuracy, especially in large-scale environments under challenging conditions. Instead, we propose to extract globally reliable features by implicitly embedding high-level semantics into both the detection and description processes. Specifically, our semantic-aware detector is able to detect keypoints from reliable regions (e.g. building, traffic lane) and suppress unreliable areas (e.g. sky, car) implicitly instead of relying on explicit semantic labels.  This boosts the accuracy of keypoint matching by reducing the number of features sensitive to appearance changes and avoiding the need of additional segmentation networks at test time. Moreover, our descriptors are augmented with semantics and have stronger discriminative ability, providing more inliers at test time. Particularly, experiments on long-term large-scale visual localization Aachen Day-Night and RobotCar-Seasons datasets demonstrate that our model outperforms previous local features and gives competitive accuracy to advanced matchers but is about 2 and 3 times faster when using 2k and 4k keypoints, respectively.  Code is available at \href{https://github.com/feixue94/sfd2}{https://github.com/feixue94/sfd2}.
\end{abstract}

\section{Introduction}

Visual localization is key to various applications including autonomous driving and robotics. Structure-based algorithms~\cite{as,csl,smc,ssm,hfnet,lbr} involving mapping and localization processes still dominate in large-scale localization. Traditionally, handcrafted features (\eg SIFT~\cite{sift, rootsift}, ORB~\cite{orb}) are widely used. However, these features are mainly based on statistics of gradients of local patches and thus are prone to appearance changes such as illumination and season variations in the long-term visual localization task. With the success of CNNs, learning-based features~\cite{d2-net,r2d2,aslfeat,superpoint,lfnet,lift,disk} are introduced to replace handcrafted ones and have achieved excellent performance. With massive data for training, these methods should be able to automatically extract keypoints from more reliable regions (\eg building, traffic lane) by focusing on discriminative features~\cite{ssm}. Nevertheless, due to the lack of explicit semantic signals for training, their ability of selecting globally reliable keypoints is limited, as shown in Fig.~\ref{fig:heatmap} (detailed analysis is provided in Sec.~B.1 in the supplementary material). Therefore, they prefer to extract locally reliable features from objects including those which are not useful for long-term localization (\eg sky, tree, car), leading to limited accuracy, as demonstrated in Table~\ref{tab:aachen}.

\begin{figure}[t]
	\centering
	\includegraphics[width=.8\linewidth]{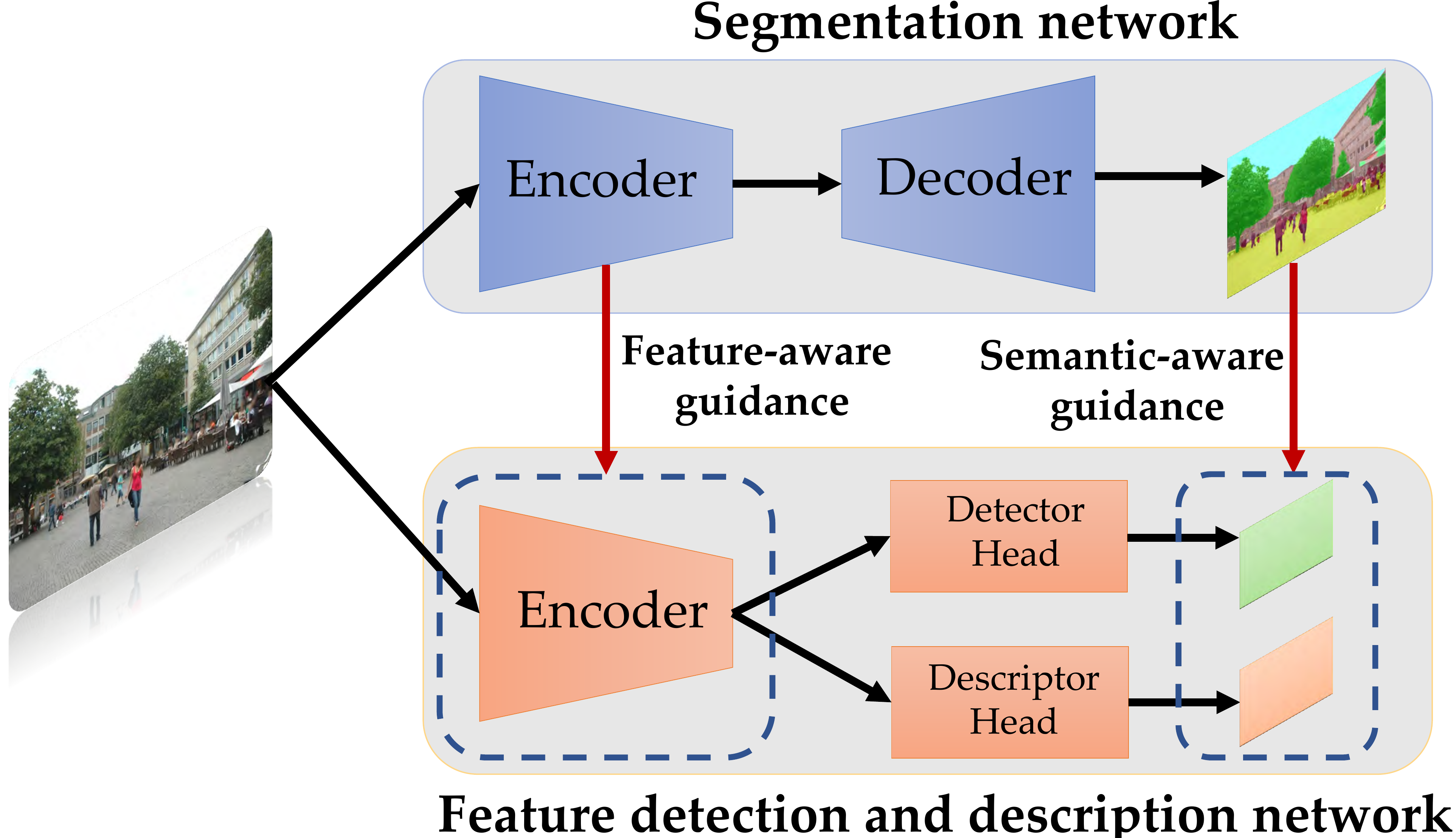}
	\caption{\textbf{Overview of our framework.} Our model implicitly incorporates semantics into the detection and description processes with guidance of an off-the-shelf segmentation network during the training process. Semantic- and feature-aware guidance are adopted to enhance its ability of embedding semantic information.}
	\label{fig:framework}
\end{figure}

Recently, advanced matchers based on sparse keypoints~\cite{superglue,sgmnet,clustergnn} or dense pixels~\cite{loftr,dualresolution,enet,pdcnet,ahm2019,s2dnet,dgcnet,aspanformer2022} are proposed to enhance keypoint/pixel-wise matching and have obtained remarkable accuracy. Yet, they have quadratic time complexity due to the attention and correlation volume computation. Moreover, advanced matchers rely on spatial connections of keypoints and perform image-wise matching as opposed to fast point-wise matching, so they take much longer time than nearest neighbor matching (NN) in both mapping and localization processes because of a large number of image pairs (much larger than the number of images)~\cite{sgmnet,imp}. Alternatively, some works leverage semantics~\cite{lbr,ssm,smc} to filter unstable features to eliminate wrong correspondences and report close even better accuracy than advanced matchers~\cite{lbr}. However, they require additional segmentation networks to provide semantic labels at test time and are fragile to segmentation errors.

Instead, we implicitly incorporate semantics into a local feature model, allowing it to extract robust features automatically from a single network in an end-to-end fashion. In the training process, as shown in Fig.~\ref{fig:framework}, we provide explicit semantics as supervision to guide the detection and description behaviors. Specifically, in the detection process, unlike most previous methods~\cite{superpoint,r2d2,d2-net,aslfeat,posfeat} adopting exhaustive detection, we employ a semantic-aware detection loss to encourage our detector to favor features from reliable objects (\eg building, traffic lane) and suppress those from unreliable objects (\eg sky). In the description process, rather than utilizing triplet loss widely used for descriptor learning~\cite{hardnet,d2-net}, we employ a semantic-aware description loss consisting of two terms: inter- and intra-class losses. The inter-class loss embeds semantics into descriptors by enforcing features with the same label to be close and those with different labels to be far. The intra-class loss, which is a soft-ranking loss~\cite{aploss}, operates on features in each class independently and differentiates these features from objects of the same label. Such use of soft-ranking loss avoids the conflict with inter-class loss and retains the diversity of features in each class (\eg features from buildings usually have larger diversity than those from traffic lights). With semantic-aware descriptor loss, our model is capable of producing descriptors with stronger discriminative ability. Benefiting from implicit semantic embedding, our method avoids using additional segmentation networks at test time and is less fragile to segmentation errors.

As the local feature network is much simpler than typical segmentation networks \eg UperNet~\cite{deeplab3plus}, we also adopt an additional feature-consistency loss on the encoder to enhance its ability of learning semantic information. To avoid using costly to obtain ground-truth labels, we train our model with outputs of an off-the-shelf segmentation network~\cite{convnet,upernet}, which has achieved SOTA performance on the scene parsing task~\cite{ade20k}, but other semantic segmentation networks (\eg~\cite{deeplab3plus}) can also be used. 

 \begin{figure}[t]
	\centering
	\includegraphics[width=1.\linewidth]{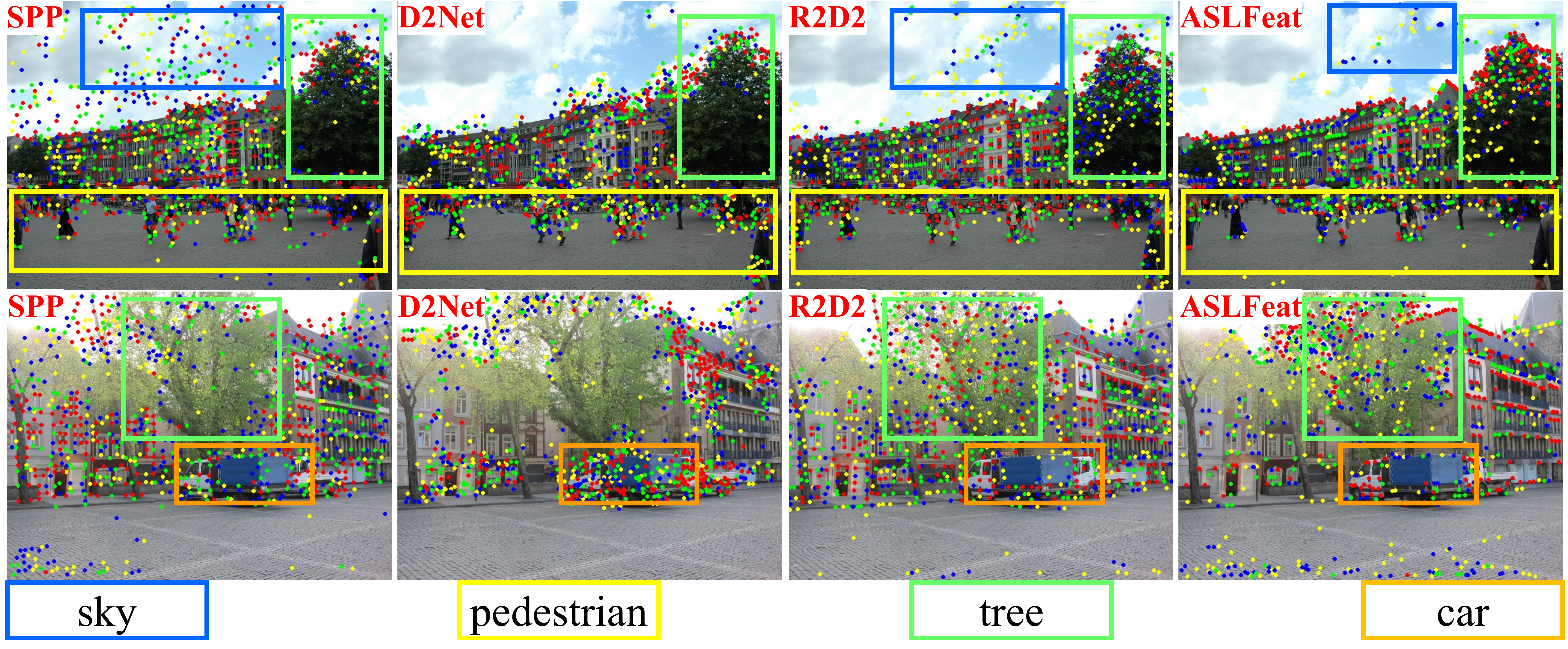}
	\caption{\textbf{Locally reliable features}. We show top 1k keypoints (reliability high$\rightarrow$low: \colorbox{red}{1-250}, \colorbox{green}{251-500}, \colorbox{blue!70}{501-750}, \colorbox{yellow}{751-1000}) of prior local features including SPP~\cite{superpoint}, D2Net~\cite{d2-net}, R2D2~\cite{r2d2}, and ASLFeat~\cite{aslfeat}. They indiscriminatively give high reliability  to patches with rich textures even from objects \eg sky, tree, pedestrian and car, which are reliable for long-term localization (best view in color). }
	\label{fig:heatmap}
\end{figure}

An overview of our system is shown in Fig.~\ref{fig:framework}. We embed semantics implicitly into the feature detection and description network via the feature-aware and semantic-aware guidance in the training process. At test time, our model produces semantic-aware features from a single network directly. We summarize contributions as follows:

\begin{itemize}
	\item We propose a novel feature network which implicitly incorporates semantics into detection and description processes at training time, enabling the model to produce semantic-aware features end-to-end at test time.
	
	\item We adopt a combination of semantic-aware and feature-aware guidance strategy to make the model embed semantic information more effectively.
	
	\item Our method outperforms previous local features on the long-term localization task and gives competitive accuracy to advanced matchers but has higher efficiency.
\end{itemize}

Experiments show our method achieves a better trade-off between accuracy and efficiency than advanced matchers~\cite{superglue,sgmnet,clustergnn} especially on devices with limited computing resources. We organize the rest of this paper as follows. In Sec.~\ref{sec:relatedworks}, we introduce related works. In Sec.~\ref{sec:method}, we describe our method in detail. We discuss experiments and limitations in Sec.~\ref{sec:experiments} and Sec.~\ref{sec:limitation} and conclude the paper in Sec.~\ref{sec:conclusion}.

\section{Related Work}
\label{sec:relatedworks}

In this section, we discuss related work on visual localization, feature extraction and matching, and knowledge distillation. 

\textbf{Visual localization.} Visual localization methods can be roughly categorized as image-based and structure-based. Image-based systems recover camera poses by finding the most similar one in the database with global features, \eg NetVLAD~\cite{netvlad}, CRN~\cite{learncontext2017}. Due to the limited number of images in the database, they can only give approximate poses. To obtain more precise poses, structure-based methods build a sparse 3D map via SfM and estimate the pose via PnP from 2D-3D correspondences~\cite{as,csl,visuallocalization,hfnet,cascadedfiltering,lbr}.  Some other works have tried to predict the camera pose directly from images, \eg PoseNet~\cite{posenet} and its variations~\cite{glnet, lsg}, or regress scene coordinates~\cite{dsac,dsac++,localinstance,vsnet}. However, the former have been proved to perform similar to image retrieval~\cite{sattler2019understanding} and latter are hard to scale to large-scale scenes~\cite{fgsn}.

\textbf{Local features.} Handcrafted features~\cite{orb,sift,rootsift} have been investigated for decades and we refer readers a survey~\cite{matching2021} for more details and focus on learned features. With the success of CNNs, learned features are proposed to replace handcrafted descriptors~\cite{l2-net,hardnet,sosnet,contextdesc,logpolardesc, spacialcoding,lisrd,crossdescriptor,affnet}, detectors~\cite{d2d,llf,sips2019}, or both~\cite{lift,superpoint,d2-net,r2d2,aslfeat,disk,posfeat}. HardNet~\cite{hardnet} focuses on metric learning by maximizing the distance between the closest positive and negative examples. Instead of using pixel-wise correspondences for training, CAPS~\cite{caps}, PoSFeat~\cite{posfeat} and PUMP~\cite{pump} utilize camera pose and local consistency of matches for supervision. SuperPoint (SPP)~\cite{superpoint} takes keypoint detection as a supervised task, training detector from synthetic geometric shapes. D2-Net~\cite{d2-net} uses local maxima across the channels as score map. R2D2~\cite{r2d2} considers both the repeatability and reliability and adopts the average precision loss~\cite{aploss} for descriptor training. ASLFeat~\cite{aslfeat} employs deformable CNNs to learn shape-aware dense features. As they focus mainly on \textit{local reliability} of features, regardless of their superior accuracy to handcrafted features, their performance is limited in the long-term large-scale localization task. To further improve the accuracy, some works~\cite{matchorno2020, learnrep2021,predmatch2014} learn to filter unstable keypoints with extra matching score, repeatability or semantic labels. Essentially different with these methods, our model detects and extracts semantic-aware features automatically in an end-to-end fashion. As a result, our features are able to produce more accurate localization results.

\textbf{Advanced matcher.} As NN matching is unable to incorporate spatial connections of keypoints for matching, advanced matchers are proposed to enhance the accuracy by leveraging the spatial context of a set of keyppoints~\cite{superglue,sgmnet,clustergnn} or an image patch~\cite{dualresolution,sparsenc,patch2pix,loftr,ahm2019,aspanformer2022}. SuperGlue (SPG)~\cite{superglue} utilizes graph neural networks with attention mechanism to propagate information among keypoints. It produces impressive accuracy, whereas its time complexity is quadratic to the number of keypoints. This problem is partially mitigated by using seeded matching~\cite{sgmnet} and cluster matching~\cite{clustergnn}, but the time is still thousands of times slower than NN matching. Dense matchers~\cite{sparsenc,dualresolution,loftr,aspanformer2022} compute pixel-wise correspondence from correlation volumes, so they undergo the high time and memory cost as sparse matchers~\cite{superglue,sgmnet,clustergnn}. Moreover, advanced matchers operates on image pairs as opposed to keypoints, so considering the number of image pairs, systems with advanced matchers could be much slower in real applications, as analyzed in ~\cite{sgmnet}. In this paper, we embed high-level semantic information into local features implicitly to enhance both feature detection and description, enabling our model with simple NN matching to yield comparable results to advanced matchers. Our work provides a good trade-off of time and accuracy especially on devices with limited computing resources.

\textbf{Visual semantic localization.} Compared to local features, high-level semantics are more robust to appearance changes and have been widely used in visual localization~\cite{fgsn,stenborg2018long,localinstance,vsnet,svl,lln,smc,ssm, dasgil,svl2017,matchfeat2014, semantics-aware2017}. LLN~\cite{lln} and SVL~\cite{svl2017} use the discriminative landmarks for place recognition. ToDayGAN~\cite{todaygan} transfers night images to day images with GAN~\cite{gan}. MFC~\cite{matchfeat2014}, SMC~\cite{smc}, SSM~\cite{ssm}, and DASGIL~\cite{dasgil} incorporate segmentation networks into a standard localization pipeline to reject semantically-inconsistent matches. More recently, LBR~\cite{lbr} learns to recognize global instances for both coarse and fine localization. In fine localization, it filters unstable features and conducts instance-wise matching, achieving close accuracy to advanced matchers~\cite{superglue}. Unlike these methods, which require additional models to provide explicit semantic labels at test time, we embed the semantic information into the network and produce semantic-aware features directly from a single network. 

\textbf{Knowledge distillation.} Knowledge distillation techniques have been widely used for tasks including model compression~\cite{hfnet} and knowledge transfer~\cite{cross-model-knowledge}. Our usage of pseudo ground-truth local reliability and semantic labels predicted by off-the-shelf networks is more like a knowledge transfer task. In this paper, we focus mainly on how to effectively leverage the high-level semantics for low-level feature extraction.

\section{Method}
\label{sec:method}

\begin{figure*}[t]
	\centering
	\includegraphics[width=1.\linewidth]{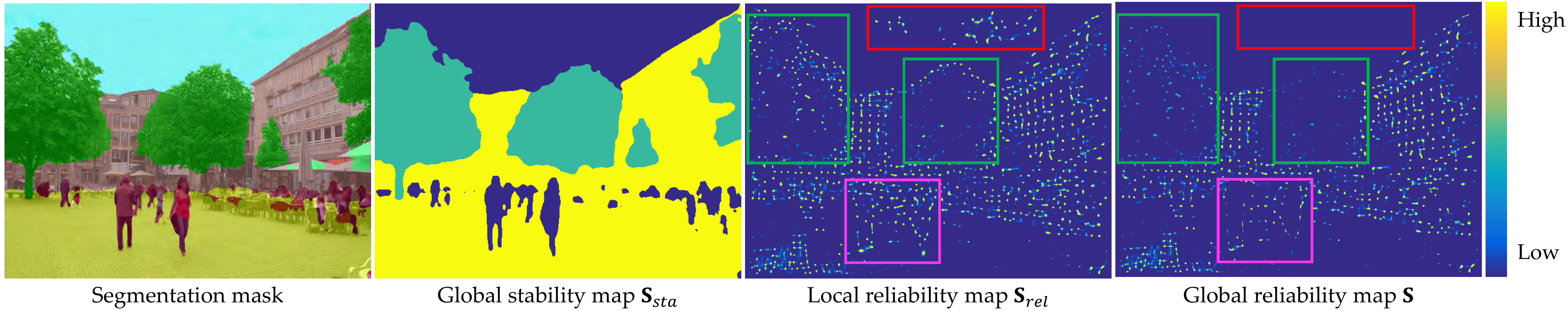}
	\caption{\textbf{Semantic-guided feature detection.} From left to right: semantic segmentation mask predicted by~\cite{convnet,upernet}, stability map $\mathbf{S}_{sta}$ generated according to Table~\ref{tab:stability}, local reliability map $\mathbf{S}_{rel}$ produced by SuperPoint~\cite{superpoint}, and the final global reliability map $\mathbf{S}$. Local reliability map gives very high score to clouds (\textcolor{red}{red}), trees (\textcolor{green}{green}), and pedestrians (\textcolor{magenta}{pink}) in addition to buildings, while the global reliability map removes unstable regions (sky, pedestrians), suppresses short-term objects (trees), and retains stable areas (buildings).}
	\label{fig:segmap}
\end{figure*}

 As shown in Fig.~\ref{fig:framework}, our model consists of an encoder $\mathcal{F}_{enc}$ and two decoders $\mathcal{F}_{det}$ and $\mathcal{F}_{desc}$. $\mathcal{F}_{enc}$ extracts high-level features $\mathbf{X}$ from image $\mathbf{I} \in R^{3 \times H \times W}$. $\mathcal{F}_{det}$ predicts the reliability map $\mathbf{S} \in R^{H \times W}$ and $\mathcal{F}_{desc}$ produces descriptors $\mathbf{X}_{desc} \in R^{128 \times \frac{H}{4} \times \frac{W}{4}}$. $H$ and $W$ are the height and width of the image. In this section, we give details about how to implicitly incorporate semantic information into our feature detection and description processes.

\subsection{Semantic-guided Feature Detection}
The detector predicts the reliability map as $\mathbf{S} = \mathcal{F}_{det}(\mathbf{X})$. Previously, the reliability map $\mathbf{S}$ is defined by the richness of textures in patches (\eg response to corners~\cite{superpoint} or blobs~\cite{sift}). Recently, learned local features~\cite{d2-net,r2d2,aslfeat,posfeat} define the reliability on the discriminative ability of descriptors. As shown in Fig.~\ref{fig:heatmap}, these two definitions, however, only reveal the reliability of pixels at a local level but lack the stability at a global level. Instead, we redefine the reliability of features by taking both the local reliability $\mathbf{S}_{rel}$ and global stability $\mathbf{S}_{sta}$ into consideration.

\textbf{Local reliability.} Local reliability shows the robustness of a keypoint to appearance changes and viewpoint variations. Previous learning-based features adopt two strategies for reliable feature learning: learning from groundtruth corners~\cite{superpoint} and learning from the discriminative ability of descriptors~\cite{d2-net,aslfeat,r2d2,posfeat}. We find that corners~\cite{superpoint} are more robust compared to purely learned detectors, as shown in~\cite{lbr,pump}, where SPP detector achieves better results when replacing other detectors. Therefore, following~\cite{hfnet}, we use the detection score $\mathbf{S}_{rel}$ of SPP~\cite{superpoint} as pseudo groundtruth, which is one of the best corner detectors. At the same time, local reliability is slightly adjusted by the discriminative ability of descriptors in the training process (see Sec.~\ref{sec:sem_desc}).

\textbf{Global stability.} The global stability of a pixel is assigned based on the semantic label which it belongs to. Specifically, we group all 120 semantic labels in ADE20k dataset~\cite{ade20k}, according to how they change over time, into four categories, denoted as \textit{Volatile}, \textit{Dynamic}, \textit{Short-term}, and \textit{Long-term} in Table~\ref{tab:stability}. Volatile objects (\eg sky, water) are constantly changing and are redundant for localization. Dynamic objects (\eg car, pedestrian) are moving everyday and could cause localization error by introducing wrong matches. Short-term objects (\eg tree) can be used for short-term localization tasks (\eg VO/SLAM), yet they are sensitive to changes of illumination (low albedo) and season conditions. Long-term objects (\eg building, traffic light) are resistant to aforementioned changes and are ideal for long-term localization.

Instead of directly \textbf{filtering} unstable features~\cite{lbr,smc,ssm}, we \textbf{rerank} features with stability values assigned empirically according to the extent of desired suppression. In detail, Long-term objects are robust for both short and long-term localization, so their stability value is set to 1.0. Short-term objects are useful for short-term localization, so we set their stability to 0.5. The stability value of Volatile and Dynamic categories is set to 0.1 as they are not useful for both short/long-term localization. Note that we set it to 0.1 rather than 0. Our reranking strategy encourages the model to use stable features preferentially and uses keypoints from other objects as compensation when insufficient stable keypoints can be found, increasing the robustness of our model to various tasks (\eg feature matching, short-term localization). Fig.~\ref{fig:segmap} shows stability map $\mathbf{S}_{sta}$ transformed from Table~\ref{tab:stability}. Our current global stability is assigned based on predefined semantic labels, but a learned one might provide better performance and deserves further exploration. 

\setlength{\tabcolsep}{2.pt}

\begin{table}[t]
		\footnotesize
	\centering
	\begin{threeparttable}
		\begin{tabular}{lccccc}
			\toprule
			Category & Volatile & Dynamic & Short-term & Long-term & Stability \\
			\midrule
			sky, water & \cmark &  &   &  & $0.1$\\
			vehicle, pedestrian &  & \cmark &  &  & $0.1$ \\
			plant, grass &  &  & \cmark &  & $0.5$ \\
			building, traffic light &  &  &  & \cmark & $1.0$ \\
			\toprule
		\end{tabular}
	\end{threeparttable}
\caption{\textbf{Stability map}. Semantic labels are categorized into four groups denoted as \textit{Volatile}, \textit{Dynamic}, \textit{Short-term}, and \textit{Long-term}. Four categories are empirically assigned with  different stability values according to their robustness to appearance changes.}
\label{tab:stability}
\end{table} 

\textbf{Semantic-guided detection.} The global reliability map $\mathbf{S}_{gt}$ is generated by multiplying the local reliability map $\mathbf{S}_{rel}$ and global stability map, as $\mathbf{S}_{gt}=\mathbf{S}_{rel} \odot \mathbf{S}_{sta}$ ($\odot$ is element-wise multiplication). Fig.~\ref{fig:segmap} shows that local reliability map gives high score for all pixels with rich textures even those from the sky, pedestrians, and trees, which are useless for localization. However, the global reliability map considering both local reliability and global stability discards these sensitive features and suppresses short-term keypoints effectively. The detection loss is defined as:
\begin{align}
	L_{det}=BCE(\mathbf{S}, \mathbf{S}_{gt}),
	\label{eq:det_loss}
\end{align}
where $BCE$ denotes the \textit{binary cross-entropy} loss.

\subsection{Semantic-guided Feature Description}
\label{sec:sem_desc}
\begin{figure}[t]
	\centering
	\includegraphics[width=.95\linewidth]{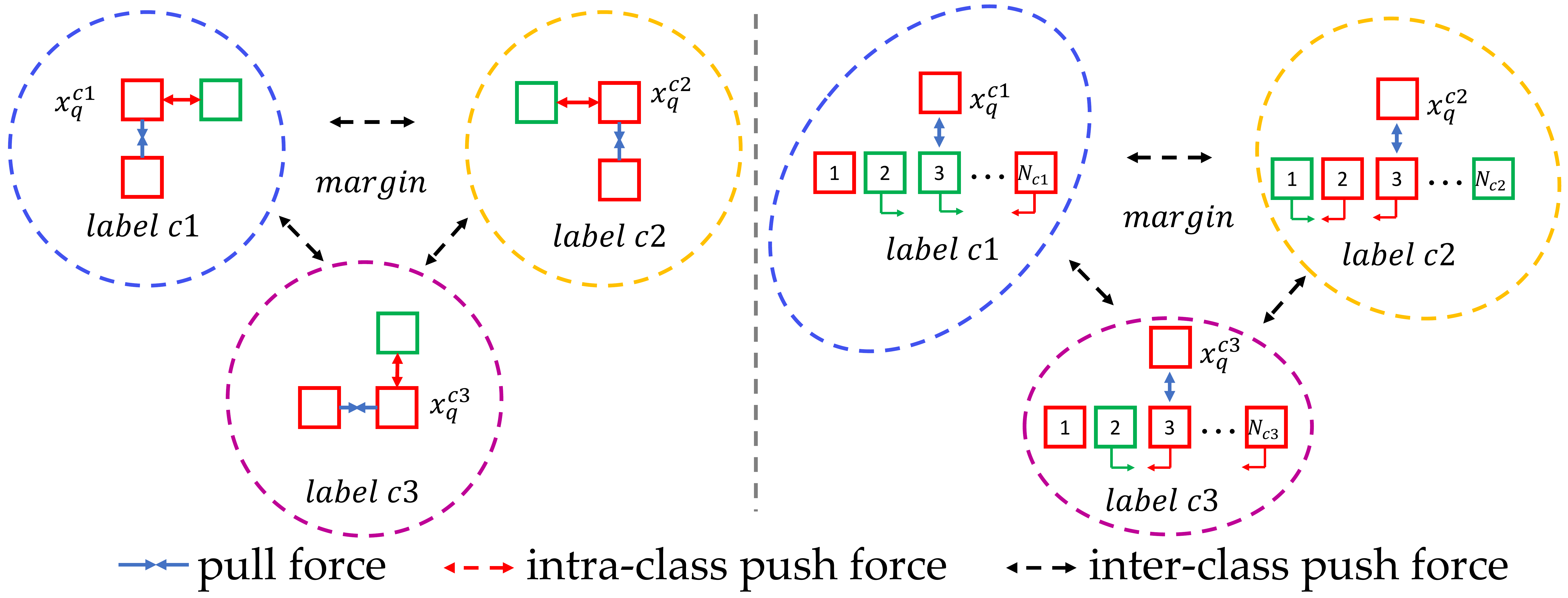}
	\caption{\textbf{Semantic-guided feature description}. Simply optimizing inter- and intra-class losses with hard margins may cause accuracy loss due to two conflicting forces (push forces of inter- and intra-class) (left). Instead, we optimize the intra-class force with a hard margin, but apply a soft ranking loss for inter-class force to avoid conflicts and retain the inner diversity of each object (right).}
	\label{fig:desc}
\end{figure}

We also enhance the discriminative ability of descriptors by embedding semantics into them directly. Unlike previous descriptors~\cite{r2d2, aslfeat, hardnet, superpoint, caps, pump},  which only differentiate keypoints based on local patch information, our descriptors enforce similarities of features in the same class while retain dissimilarities for intra-class matching. However, the two forces conflict with each other during the training process, because class-level discriminative ability needs to squeeze the space of descriptors in the same class and intra-class discriminative ability has to increase the space. A simple solution could be to set a hard margin for all classes (Fig.~\ref{fig:desc} left), but it would lead to the loss of objects' inner diversity (\eg, almost all traffic lights are similar but different buildings vary dramatically), which is indispensable for intra-class matching. To solve this problem, we design the inter-class and intra-class losses based on two different metrics (Fig.~\ref{fig:desc} right). 

\textbf{Inter-class loss.} We first enforce the semantic consistency of features by maximizing the Euclidean distance between descriptors with different labels. This allows features to find correspondences from candidates with the same labels, reducing the search space and thus improving the matching accuracy. We define the inter-class loss based on triplet loss with a hard margin to separate all possible positive and negative keypoints with different labels in a batch: 

\begin{footnotesize}
	\begin{align}
		L_{inter} = \frac{1}{N}\sum(||\mathbf{x}_i^{c_1} - \mathbf{x}_j^{c_1}||_2 - ||\mathbf{x}_i^{c_1} - \mathbf{x}_k^{c2}||_2 + m), 
		\label{eq:inter_loss}
	\end{align}
\end{footnotesize}
where $\mathbf{x}_i^{c_1}, \mathbf{x}_j^{c_1}$, and $\mathbf{x}_k^{c_2}$ are vectorized descriptors with labels of $c_1, c_1$, and $c_2$ ($c_1 \neq c_2$). $m$ is the margin and set to $1.0$. This loss is conducted on features in the whole batch and $N$ is the total number of features in a batch.

\textbf{Intra-class loss.} To make sure that the intra-class loss doesn't conflict with the inter-class loss, we relax the limitation of distances between descriptors with the same label. Instead of using triplet loss with hard margins, we adopt a soft ranking loss~\cite{aploss} by optimizing the rank of positive and negative samples rather than their distances. We use the same strategy as~\cite{r2d2} to generate positive and negative samples for each feature $\mathbf{x}^{c}_i$ from self and the other images respectively, but enforce both positive and negative samples to share the same class label $c$ as $\mathbf{x}^{c}_i$. By optimizing ranks of all samples rather than forcing a hard boundary between positive and negative pairs as the triplet loss with a hard margin does, the soft ranking loss also retains the diversity of features on objects in the same class, as shown in Fig.~\ref{fig:desc} (right). The ranking loss is based on the averaging precision (AP) loss~\cite{r2d2, aploss}:

\begin{footnotesize}
\begin{align}
	L_{intra} = \frac{1}{C}\sum_{c=1}^{C}\frac{1}{N_c}\sum_{i=1}^{N_c}(1 - AP(\mathbf{x}^{c}_i, \mathbf{S}^{\mathbf{x}^{c}_i})),
	\label{eq:intra_loss}
\end{align}
\end{footnotesize}
where $\mathbf{x}^{c}_i$ and $\mathbf{S}^{\mathbf{x}^{c}_i}$ are the query descriptor with label $c$ and corresponding predicted local reliability. $C$ and $N_c$ are the total number of classes and features in class $c$. Note that the AP loss for sample $\mathbf{x}^{c}_i$ is weighted by its reliability  $\mathbf{S}^{\mathbf{x}^{c}_i}$. Our final descriptor loss $L_{desc}$ is the combination of $L_{inter}$ and $L_{intra}$, balanced by $w_{inter}$ and $w_{intra}$:
\begin{align}
	L_{desc} = w_{inter}L_{inter} + w_{intra}L_{intra}.
	\label{eq:desc_loss}
\end{align}

\begin{figure}[t]
	\centering
	\includegraphics[width=.95\linewidth]{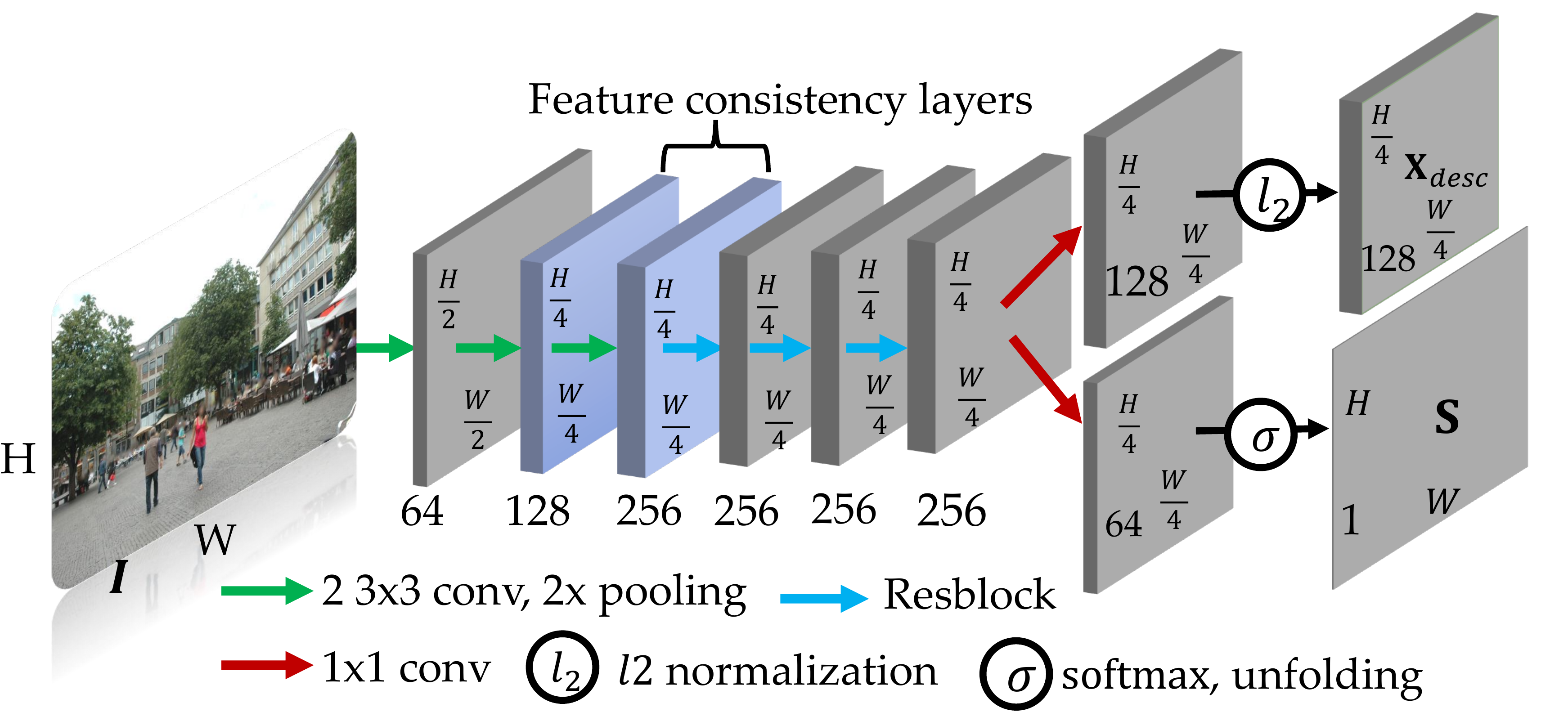}
	\caption{\textbf{Architecture of our network.} Features are 4$\times$ downsampling to save time and memory cost. Resblocks~\cite{resnet} are adopted to enhance the model's capacity. We enforce the consistency between outputs of the middle two layers of our encoder and features of the segmentation network to enhance the ability of our model to embed semantic information during the training process.}
	\label{fig:network}
\end{figure}

\subsection{Implicit Feature Guidance}
With semantic-aware detection and description losses, our model is able to learn semantic-aware features. However, compared with feature learning, semantic prediction is a more complicated task, requiring powerful encoders and aggregation layers~\cite{segnet2017, convnet} for semantic-aware feature embedding. To further improve the ability of our model to embed semantic information, we take inspiration from current knowledge distillation tasks~\cite{gou2021knowledge} and introduce a feature consistency loss on intermediate outputs of the first three layers of the encoder.

Fig.~\ref{fig:network} shows the architecture of our network. We take intermediate outputs of the encoder of ConvNeXt~\cite{convnet} as supervision signal and enforce $l_1$ loss on features of the middle 2 layers of our model:
\begin{align}
	L_{feat} = \frac{1}{2}\sum_{i=1}^{2}|\mathbf{X}_i- \mathbf{X}_i^{ConvNeXt}|,
\end{align}
where $\mathbf{X}_i$ and $\mathbf{X}_i^{ConvNeXt}$ are features of the $i$th layer in our model and ConvNeXt~\cite{convnet}, respectively. The total loss $L_{total}$ is the combination of detection loss $L_{det}$, description loss $L_{desc}$, and feature consistency loss $L_{feat}$ with weights of $w_{det}$, $w_{desc}$, and $w_{feat}$:
\begin{align}
	L_{total} = w_{det}L_{det} + w_{desc}L_{desc} + w_{feat}L_{feat}.
\end{align}

\section{Experiments}
\label{sec:experiments}
We first give implementation details. Then, we test our method on visual localization tasks in Sec.~\ref{sec:result_localization} and analyze the running time in Sec.~\ref{sec:exp_time}. Finally, we perform an ablation study in Sec.~\ref{sec:ablation_study}. More implementation details, results and analysis are provided in the \textbf{supplementary material}.

\textbf{Implementation.} We use the identical training dataset as R2D2~\cite{r2d2}. The training dataset consists of reference images in Aachen\_v1.0 dataset~\cite{aachen} and web images. As R2D2~\cite{r2d2} and LBR~\cite{lbr}, training images are augmented with style transfer. To mitigate segmentation uncertainties caused by style transfer, semantic labels of stylized images are generated from their corresponding normal images. The network is implemented in PyTorch~\cite{pytorch} and trained using Adam~\cite{adam} optimizer with $\beta_1=0.9, \beta_2=0.99$, batch size of 4, weight decay of $4 \times 10^{-4}$ on a single 3090Ti GPU for 40 epochs. The hyper-parameter $w_{intra}$ is set to 0.5, while $w_{inter}$, $w_{det}$, $w_{desc}$, and $w_{feat}$ are set to 1.0. 

\setlength{\tabcolsep}{4.pt}

\begin{table}
	\scriptsize
	\centering
		\begin{tabular}{llcc}
			\toprule
			Group & Method & Day & Night\\
			& & \multicolumn{2}{c}{$(2^\circ,0.25m)/ (5^\circ,0.5m)/ (10^\circ,5m)$} \\	
			\midrule
			
			\multirow{3}{*}{C} & AS\_v1.1~\cite{as} & {\color{red}85.3} / {\color{red}92.2} / {\color{red}97.9} & {\color{red}39.8} / {\color{red}49.0} / {\color{red}64.3} \\
			& CSL~\cite{csl} & 52.3 / 80.0 / 94.3 & 29.6 / 40.8 / 56.1 \\
			& CPF~\cite{cascadedfiltering} &  76.7 / 88.6 / 95.8 & 33.7 / 48.0 / 62.2  \\
			& \textbf{Ours} & \textbf{88.2} / \textbf{96.0} / \textbf{98.7} & \textbf{87.8} / \textbf{94.9} / \textbf{100.0} \\
			\midrule
			
			\multirow{4}{*}{S} & SSM~\cite{ssm} & 71.8 / 91.5 / 96.8 & 58.2 / 76.5 / 90.8 \\
			& VLM~\cite{lln} & 62.4 / 71.8 / 79.9 & 35.7 / 44.9 / 54.1 \\
			& SMC~\cite{smc} & 	52.3 / 80.0 / 94.3 & 29.6 / 40.8 / 56.1\\
			& LBR~\cite{lbr} & \textbf{88.3} / {\color{red}95.6} / \textbf{98.8} & {\color{red}84.7} / {\color{red}93.9} / \textbf{100.0}  \\
			& \textbf{Ours} & {\color{red}88.2} / \textbf{96.0} / {\color{red}98.7} & \textbf{87.8} / \textbf{94.9} / \textbf{100.0} \\
			\midrule
			
			\multirow{9}{*}{L} & SIFT ~\cite{sift} & 82.8 / 88.1 / 93.1 & 30.6 / 43.9 / 58.2 \\
			& SPP~\cite{superpoint} & 80.5 / 87.4 / 94.2 & 42.9 / 62.2 / 76.5 \\
			& D2Net~\cite{d2-net} & {\color{red}84.8} / {\color{red}92.6} / {\color{red}97.5} & {\color{red}84.7} / {\color{red}90.8} / 96.9  \\
			& R2D2~\cite{r2d2} &  & 76.5 / {\color{red}90.8} / \textbf{100.0}  \\
			& SIFT+CAPS~\cite{sift,caps} &  & 77.6 / 86.7 / {\color{red}99.0}  \\
			& SPP+CAPS~\cite{superpoint,caps} & & 82.7 / 87.8 / \textbf{100.0}   \\
			& SPP+LISRD~\cite{lisrd,superpoint} &  & 78.6 / 86.7 / 98.0 \\
			& SPP+PUMP~\cite{pump, superpoint} & & 74.4 / 88.0 / 98.4 \\
			& R2D2+PUMP~\cite{pump,superpoint} & & 73.3 / 86.9 / 98.4 \\
			& R2D2+LLF~\cite{llf, superpoint} &  & 72.4 / {\color{red}90.8} / {\color{red}99.0}\\
			& SOSNet+D2D~\cite{sosnet, d2d} & & 73.5 / 83.7 / 96.9 \\
			& PoSFeat~\cite{posfeat} & & 81.6 / {\color{red}90.8} / \textbf{100.0} \\
			& ASLFeat~\cite{aslfeat} & & 81.6 / 87.8 / \textbf{100.0} \\
			& \textbf{Ours} & \textbf{88.2} / \textbf{96.0} / \textbf{98.7} & \textbf{87.8} / \textbf{94.9} / \textbf{100.0} \\
			
			\midrule
			\multirow{7}{*}{M} 
			& ENCNet~\cite{sparsenc} & & 76.5 / 84.7 / 98.0 \\
			& Dual-RCNet~\cite{dualresolution} & & 79.6 / 88.8 / \textbf{100.0} \\
			& PDCNet~\cite{pdcnet}&  & 80.6 / 87.8 / \textbf{100.0}\\
			& DGCNet~\cite{dgcnet}& 22.9 / 49.8 / 84.7 & 14.3 / 37.8 / 79.6 \\
			& Pixloc~\cite{backtofeature}& 84.7 / 94.2 / \textbf{98.8} & 81.6 / {\color{red}93.9} / \textbf{100.0} \\
			& AHM~\cite{ahm2019} & 47.8 / 72.2 / 91.3 & 30.6 / 53.1 / 78.6  \\
			& S2DNet~\cite{s2dnet} &  84.5 / 90.3 / 95.3 & 74.5 / 82.7 / 94.9 \\
			& Patch2Pix~\cite{patch2pix} & 84.6 / 92.1 / 96.5 & 82.7 / 92.9 / {\color{red}99.0} \\
			
			& SPP+SPG~\cite{superpoint, superglue} &\textbf{89.6} / 95.4 / \textbf{98.8} & {\color{red}86.7} / {\color{red}93.9} / \textbf{100.0} \\
			& SPP+SGMNet~\cite{superpoint,sgmnet} &  86.8 / 94.2 / 97.7 & 83.7 / 91.8 / {\color{red}99.0}\\
			& SPP+ClusterGNN~\cite{superpoint, clustergnn} & {\color{red}89.4} / {\color{red}95.5} / 98.5 & 81.6 / {\color{red}93.9} / \textbf{100.0} \\
			& \textbf{Ours} & 88.2 / \textbf{96.0} / {\color{red}98.7} & \textbf{87.8} / \textbf{94.9} / \textbf{100.0} \\
			
			\bottomrule			   
		\end{tabular}
	\caption{\textbf{Results on Aachen dataset~\cite{aachen,visbenchmark}.} The best and second best results are highlighted with \textbf{bold} and {\color{red} red} fonts.}
	\label{tab:aachen}
\end{table}

\setlength{\tabcolsep}{4.pt}

\begin{table}
	\scriptsize
	\centering
		\begin{tabular}{llcc}
			\toprule
			Group & Method & Day & Night \\	
			& & \multicolumn{2}{c}{$(2^\circ,0.25m)/ (5^\circ,0.5m)/ (10^\circ,5m)$} \\	
			\midrule
			S & LBR~\cite{lbr} & \textbf{89.1} / \textbf{96.1} / \textbf{99.3} &{\color{red}77.0} / {\color{red}90.1} / \textbf{ 99.5} \\
			& \textbf{Ours} & {\color{red}88.2} / {\color{red}96.0} / {\color{red}98.7} & \textbf{78.0} / \textbf{92.1} / \textbf{99.5} \\
			
			\midrule
			
			\multirow{7}{*}{H} & SIFT~\cite{sift} & 72.2 / 78.4 / 81.7 &	19.4 / 23.0 / 27.2 \\
			&SPP~\cite{superpoint}  & 87.9 / 93.6 / 96.8 & 70.2 / 84.8 / 93.7\\
			& D2Net~\cite{d2-net} & 84.1 / 91.0 / 95.5 & 63.4 / 83.8 / 92.1 \\
			& R2D2~\cite{r2d2}& \textbf{88.8} / 95.3 / 97.8 & 72.3 / {\color{red}88.5} / 94.2 \\
			& ASLFeat~\cite{aslfeat} & 	88.0 / {\color{red}95.4} / {\color{red}98.2} & 70.7 / 84.3 / 94.2 \\
			& CAPS+SIFT~\cite{caps, sift} & 82.4 / 91.3 / 95.9 & 61.3 / 83.8 / 95.3 \\
			& LISRD+SPP~\cite{lisrd,superpoint} &  & 73.3 / 86.9 / 97.9 \\
			& LLF+R2D2~\cite{llf, superpoint}&  & 71.2 / 81.2 / 94.2\\
			& PoSFeat~\cite{posfeat} & & {\color{red}73.8} / 87.4 / {\color{red}98.4} \\
			& \textbf{Ours} & {\color{red}88.2} / \textbf{96.0} / \textbf{98.7} & \textbf{78.0} / \textbf{92.1} / \textbf{99.5} \\
			\midrule
			
			\multirow{3}{*}{M} 
			& SPP+SGMNet~\cite{superpoint, sgmnet} & 88.7 / \textbf{96.2} / 98.9 & 75.9 / 89.0 / 99.0 \\
			& SPP+SPG~\cite{superpoint, superglue} & \textbf{89.8} / {\color{red}96.1} / \textbf{99.4} & 77.0 / 90.6 / \textbf{100.0}\\
			& Patch2Pix~\cite{patch2pix} & 86.4 / 93.0 / 97.5 & 72.3 / 88.5 / 97.9\\
			& LoFTER~\cite{loftr} & 88.7 / 95.6 / {\color{red}99.0} & \textbf{78.5} / 90.6 / 99.0 \\
			& ASpanFormer~\cite{aspanformer2022} & {\color{red}89.4} / 95.6 / {\color{red}99.0} & 77.5 / {\color{red}91.6} / {\color{red}99.5} \\
			& \textbf{Ours} & 88.2 / 96.0 / 98.7 & {\color{red}78.0} / \textbf{92.1} / {\color{red}99.5} \\
			
			\bottomrule
		\end{tabular}
\caption{\textbf{Results on  Aachen\_v1.1 dataset~\cite{aachen,visbenchmark}.} The best and second best results are highlighted with \textbf{bold} and {\color{red} red} fonts.}
\label{tab:aachenv11}
\end{table}

\subsection{Long-term Large-scale Localization}
\label{sec:result_localization}
We test our method on Aachen (v1.0 and v1.1)~\cite{aachen,visbenchmark} and RobotCar-Seasons (RoCaS)~\cite{robotcar,aachen} datasets under various illumination, season, and weather conditions. Aachen\_v1.0 contains 4,328 reference and 922 (824 day, 98 night) query images captured around the Aachen city center. Aachen\_v1.1 expands v1.0 by adding 2,369 reference and 93 night query images. RoCaS has 26,121 reference and 11,934 query images. It is challenging because of various conditions of day query images (rain, snow, dusk, winter) and poor lighting of night query images in suburban areas. We adopt the success ratio at error thresholds of $(2^\circ,0.25m), (5^\circ,0.5m), (10^\circ,5m)$ as metric. We additionally provide results on Extended CMU-Seasons dataset in the \textbf{supplementary material}.

\textbf{Baselines.} Baselines include classic systems (C) \eg AS\_v1.1~\cite{as}, CSL~\cite{csl}, and CPF~\cite{cascadedfiltering} and methods using semantics (S), \eg LLN~\cite{lln}, SMC~\cite{smc}, SSM~\cite{ssm}, DASGIL~\cite{dasgil}, ToDayGAN~\cite{todaygan} and LBR~\cite{lbr}. We also compare our model with learned features~\cite{superpoint,d2-net,r2d2,caps,aslfeat,pump} (L). As prior methods~\cite{superpoint, d2-net, r2d2,caps,aslfeat,pump}, we use HLoc~\cite{hfnet} pipeline for reconstruction and mutual nearest matching (MNN). NetVLAD~\cite{netvlad} is used to offer 50 and 20 candidates for Aachen and RoCaS datasets, respectively. We additionally compare our approach with advanced sparse/dense matchers (M) \eg, Superglue (SPG)~\cite{superglue}, SGMNet~\cite{sgmnet}, ClusterGNN~\cite{clustergnn} and ASpanFormer~\cite{aspanformer2022}, LoFTER~\cite{loftr}, Patch2Pix~\cite{patch2pix}, Dual-RCNet~\cite{dualresolution}. Their results are obtained from the visual benchmark\footnote{https://www.visuallocalization.net/benchmark/} or original papers. 

\setlength{\tabcolsep}{2.5pt}

\begin{table}
	\scriptsize
	\centering
	\begin{tabular}{llccc}
		\toprule
		Group & Method & day & night & night-rain \\	
		& & \multicolumn{3}{c}{$(2^\circ,0.25m)/ (5^\circ,0.5m)/ (10^\circ,5m)$} \\	
		\midrule
		
		\multirow{3}{*}{C} & AS~\cite{as} & 43.6 / 76.0 / 94.0 & 1.6 / 3.9 / 10.5 & 2.0 / 10.9 / 18.0 \\
		& CSL~\cite{csl} & 45.3 / 73.5 / 90.1 & 0.2 / 0.9 / 5.3 & 0.9 / 4.3 / 9.1\\
		& CPF~\cite{cascadedfiltering}  & {\color{red}48.0} / {\color{red}78.0} / {\color{red}94.2} &  {\color{red}2.3} / {\color{red}6.6} / {\color{red}15.3} & {\color{red}4.5} / {\color{red}12.3} / {\color{red}18.6}\\
		& \textbf{Ours} & \textbf{56.9} / \textbf{81.6} / \textbf{97.4} & \textbf{27.6} / \textbf{66.2} / \textbf{90.2} & \textbf{43.0} / \textbf{71.1} / \textbf{90.0} \\
		\midrule
		
		\multirow{5}{*}{S} & SSM~\cite{ssm} & 54.5 / {\color{red}81.6} / 96.7 &  10.0 / 23.7 / 45.4 & 14.5 / 33.2 / 47.5\\
		& VLM~\cite{lln} & 7.9 / 30.0 / 85.9 &  11.9 / 26.0 / 55.0 & 15.7 / 34.5 / 60.5\\
		& SMC~\cite{smc}  & 50.3 / 79.3 / 95.2 &  6.2 / 18.5 / 44.3 & 8.0 / 26.4 / 46.4\\
		& DASGIL-FD~\cite{dasgil} & 8.7 / 30.7 / 81.3 &  1.6 / 4.8 / 19.9 & 1.8 / 4.3 / 21.6\\
		& ToDayGAN~\cite{todaygan, d2-net} & 52.2 / 80.1 / 95.9 & 16.4 / 43.2 / 73.3 & 24.1 / 50.5 / {\color{red}74.1} \\
		& LBR~\cite{lbr} & {\color{red}56.7} / \textbf{81.7} / \textbf{98.2} &  {\color{red}24.9} / {\color{red}62.3} / {\color{red}86.1}	& \textbf{47.5} / \textbf{73.4} / \textbf{90.0} \\ 
		& \textbf{Ours} & \textbf{56.9} / {\color{red}81.6} / \textbf{97.4} & \textbf{27.6} / \textbf{66.2} / \textbf{90.2} & {\color{red}43.0} / {\color{red}71.1} / \textbf{90.0} \\
		\midrule
		
		\multirow{6}{*}{L} & SIFT~\cite{sift} & 53.5 / 77.6 / 92.6 &   7.8 / 13.9 / 22.1 & 9.5 / 14.5 / 17.0\\
		& SPP~\cite{superpoint} & 56.5 / 81.5 / 97.1 &   16.9 / 41.6 / 71.5 &	22.0 / 45.0 / 68.0\\
		& D2Net~\cite{d2-net} & 54.5 / 80.0 / 95.3 &  18.0 / 39.7 / 53.9 & 22.7 / 40.5 / 56.1 \\
		& R2D2~\cite{r2d2} & \textbf{57.4} / \textbf{81.9} / {\color{red}97.9} &   18.3 / 43.4 / 67.8 & 29.1 / 50.2 / 68.2 \\
		& CAPS~\cite{caps} & 56.0 / 81.5 / 96.5 & 21.9 / 54.3 / {\color{red}86.8} & 27.0 / 58.9 / 85.9 \\
		& ASLFeat~\cite{aslfeat} & {\color{red}57.1} / \textbf{81.9} / \textbf{98.4} & {\color{red}23.5} / {\color{red}55.9} / 80.1 & {\color{red}41.1} / {\color{red}66.8} / {\color{red}86.1} \\
		& \textbf{Ours} & 56.9 / {\color{red}81.6} / 97.4 & \textbf{27.6} / \textbf{66.2} / \textbf{90.2} & \textbf{43.0} / \textbf{71.1} / \textbf{90.0} \\
		\midrule

		\multirow{3}{*}{M} & SPP+SPG~\cite{superpoint, superglue} & \textbf{56.9} / {\color{red}81.7} / \textbf{98.1} &   {\color{red}24.2} / 62.6 / 87.4 & 42.3 / 69.3 / 90.2\\
		& Pixloc~\cite{backtofeature} &\textbf{56.9} / \textbf{82.0} / \textbf{98.1} &   {\color{red}24.2} / {\color{red}62.8} / 88.4 & \textbf{45.5} / \textbf{72.5} / {\color{red}90.7}\\
		& AHM~\cite{ahm2019} & 45.7 / 78.0 / 95.1 &  16.2 / 55.3 / \textbf{93.6} & 28.4 / 68.4 / \textbf{95.5} \\
		& \textbf{Ours} & \textbf{56.9} / 81.6 / {\color{red}97.4} & \textbf{27.6} / \textbf{66.2} / {\color{red}90.2} & {\color{red}43.0} / {\color{red}71.1} / 90.0 \\				
	\bottomrule
	\end{tabular}
\caption{\textbf{Results on RobotCar-Seasons dataset~\cite{robotcar,visbenchmark}.} The best and second best results are highlighted with \textbf{bold} and {\color{red} red} fonts.}
\label{tab:robotcar}
\end{table}

\textbf{Comparison with classic methods (C).} As shown in Table~\ref{tab:aachen} and~\ref{tab:robotcar}, our model outperforms all classic methods. As most these methods use SIFT~\cite{sift}, they are more sensitive to weather and illumination changes than learned features.

\textbf{Comparison with methods using explicit semantics (S).} By leveraging semantic labels to filter potentially wrong matches, these models achieve better performance for day and night images in Table~\ref{tab:aachen} and~\ref{tab:robotcar} but require segmentation results at test time. Our model outperforms all other approaches (except LBR~\cite{lbr}). LBR~\cite{lbr} reports excellent accuracy by selecting keypoints from buildings and performing instance-wise matching. Our method gives close results to LBR on day images but better performance on most night images, because our model does not require explicit semantic labels and is less fragile to segmentation errors especially for night images. LBR performs better than ours on night-train images in Table~\ref{tab:robotcar} because it is trained on augmented night rainy images, while our model is trained only on generated night images as R2D2.

\textbf{Comparison with learned features (L).} Benefiting from training with massive data, learned features such as R2D2~\cite{r2d2}, ASLFeat~\cite{aslfeat} and PoSFeat~\cite{posfeat}, outperform SIFT~\cite{sift,rootsift}. As they extract keypoints indiscriminately, they are still more sensitive to appearance changes especially on night images than semantic-aware methods~\cite{lbr}, as shown in Table~\ref{tab:aachen} and \ref{tab:robotcar}. Our model extracts semantic-aware features directly, so it gives higher accuracy.

\textbf{Comparison with advanced matchers (M).} In Table~\ref{tab:aachen}, \ref{tab:aachenv11} and~\ref{tab:robotcar}, we also show the results of previous advanced matchers. We find that our approach outperforms the recent efficient variations of SPG~\cite{superglue} (\eg SGMNet~\cite{sgmnet}, ClusterGNN~\cite{clustergnn}) and gives competitive results to SPG, which achieves the best accuracy and is also the slowest method. Note that our model only uses the simple MNN for matching.  We provide a detailed analysis of time and memory usage in Sec.~\ref{sec:exp_time} and argue that our method provides a good trade off between running time and accuracy especially on devices with limited computing resources.

\begin{figure}[t]
	\centering
	\includegraphics[width=.85\linewidth]{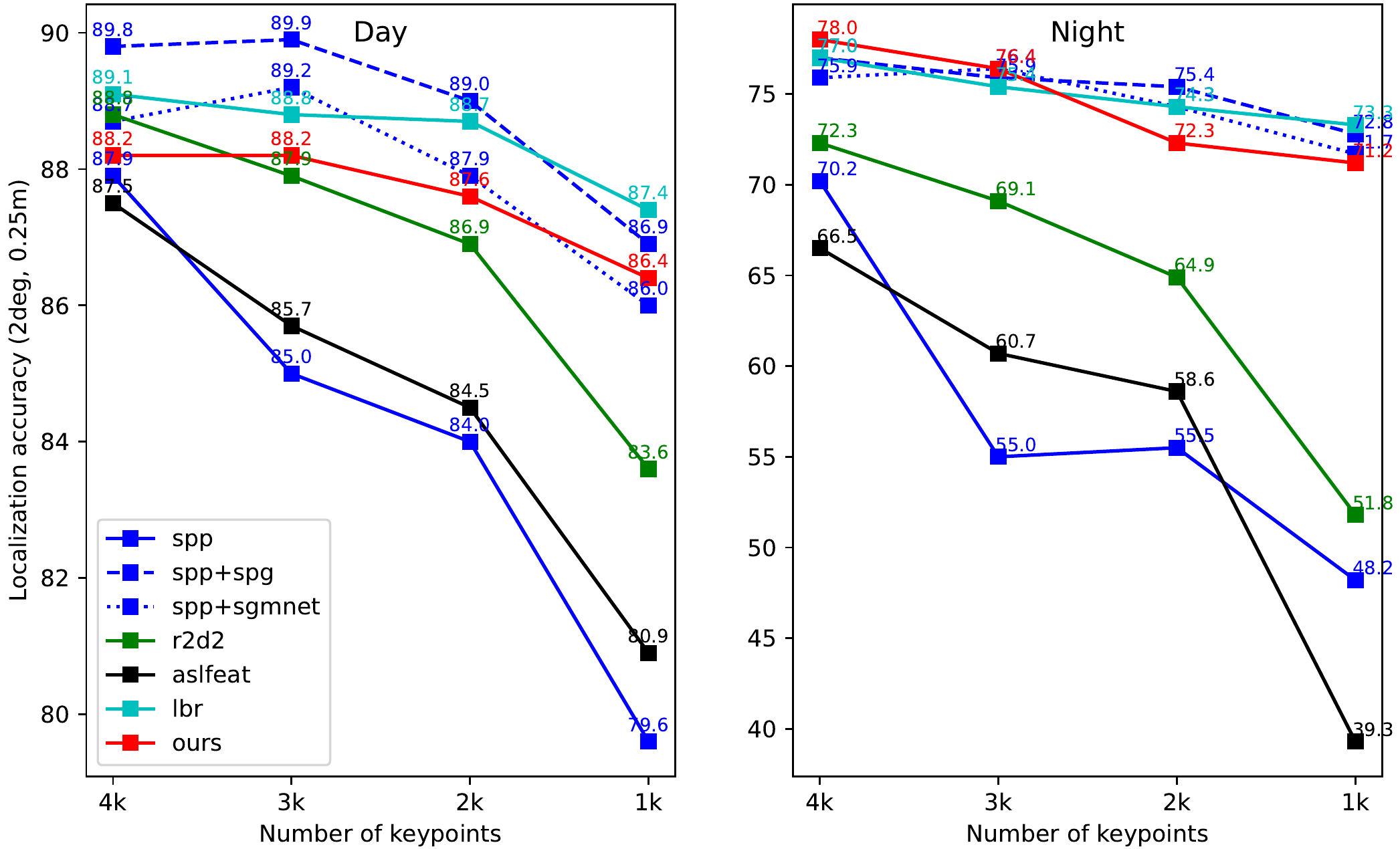}
	\caption{\textbf{Influence of the number of keypoints.} We report results of different number (from 4k to 1k) of keypoints on Aachen\_v1.1~\cite{aachen,visbenchmark} at error threshold of $(2^\circ, 0.25m)$.}
	\label{fig:num_kpts}
\end{figure}

\begin{figure*}[t]
	\centering
	\includegraphics[width=1.\linewidth]{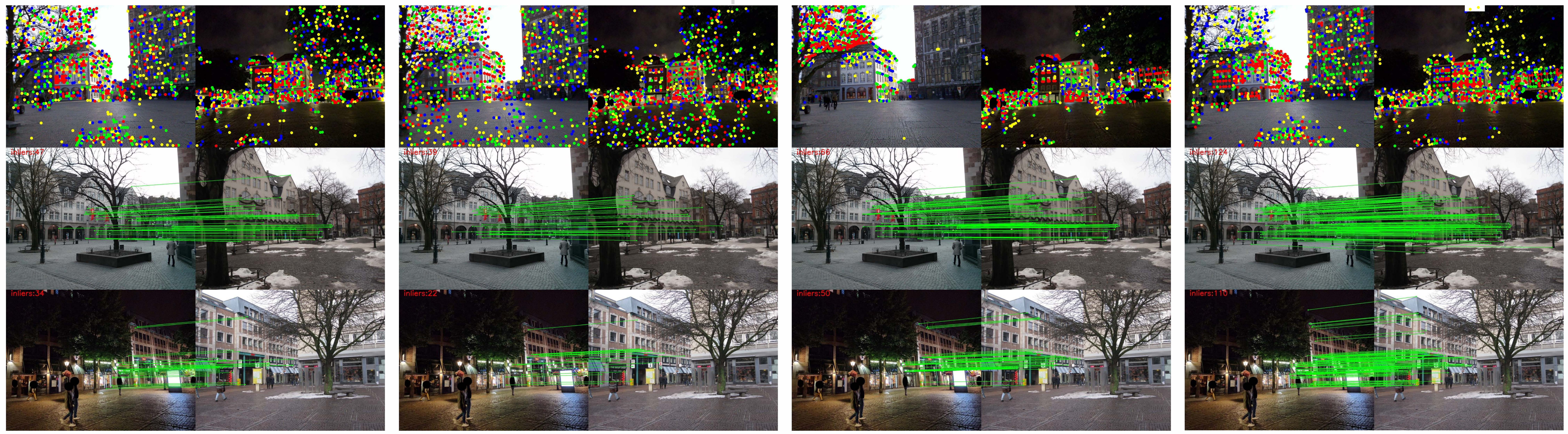}
	\caption{\textbf{Qualitative comparison of detection and matching.} Left$\rightarrow$right: SPP~\cite{superpoint}, R2D2~\cite{r2d2}, ASLFeat~\cite{aslfeat} and our method. Our model favors keypoints on stable areas (reliability high$\rightarrow$low: \colorbox{red}{1-250}, \colorbox{green}{251-500}, \colorbox{blue!70}{501-750}, \colorbox{yellow}{751-1000}) and gives more inliers.}
	\label{fig:macthes}
\end{figure*}

\textbf{Robustness to the number of keypoints}.
We observe that most previous methods~\cite{r2d2,d2-net,aslfeat,posfeat,pump} extract keypoints with the number ranging from 10k (R2D2, ASLFeat) to 40k (PosFeat) for evaluation. Although increasing the number improves the accuracy, it causes the time cost as well, which should be taken into consideration especially on devices with limited computing resources. In this experiment, we test the ability of previous and our methods of extracting fewer but more robust features by reducing the number of keypoints from 4k to 1k. Note that results of~\cite{r2d2,superpoint,superglue,lbr} are from LBR~\cite{lbr} and results of ASLFeat~\cite{aslfeat} are obtained by running official source code.

Fig.~\ref{fig:num_kpts} shows that as the number of keypoints decreases, all previous features~\cite{superpoint,r2d2,aslfeat} undergo dramatic accuracy loss especially for night images. SPP+SPG and LBR perform more robust because of the global context or semantics. With implicitly embedded semantics, our feature outperforms R2D2, ASLFeat, and SPP especially on night images and gives competitive results to SPP+SPG and LBR. 

\textbf{Qualitative comparison.} Fig.~\ref{fig:macthes} shows the detection and matching results of query images under conditions of large illumination and season changes. Compared with prior features~\cite{superpoint, r2d2, aslfeat}, which prefer keypoints from areas with rich textures, our method favors keypoints from objects robust for long-term localization (\eg buildings). When insufficient keypoints can be found from stale regions, our model also uses keypoints from Short-term objects \eg trees from compensation but assigns them with lower reliability. Besides, our feature gives more inliers for query images with large occlusions of trees and severe illumination changes. 

\subsection{Running time analysis}
\label{sec:exp_time}
\setlength{\tabcolsep}{10.pt}

\begin{table}
	\scriptsize
	\centering
	\begin{tabular}{lcc}
		\toprule
		Model & Input size & Running time (ms) \\
		\midrule
		SPP~\cite{superpoint} & 1024$\times$1024 & 13.1\\
		R2D2~\cite{r2d2} & 1024$\times$1024 & 72.4 \\
		ASLFeat*~\cite{aslfeat} & 1024$\times$1024 & 112.3 \\
		LBR (feature)~\cite{lbr} & 1024$\times$1024 & 30.1  \\
		LBR (segmentation)~\cite{lbr} & 256$\times$256 & 9.2 \\
		SPG~\cite{superglue} & 2k$\times$2k, 4k$\times$4k & 52.2, 146.5  \\
		SGMNet~\cite{sgmnet} & 2k$\times$2k, 4k$\times$4k & 85.5, 97.6 \\
		\textbf{Ours} & 1024$\times$1024 & 33.2\\
		
		\bottomrule 
	\end{tabular}
\caption{\textbf{Running time.} We report test time of prior features~\cite{superpoint,r2d2,aslfeat}, matchers~\cite{superglue,sgmnet} and our method (*in TensorFlow).}
\label{tab:time_memory}
\end{table}
Table~\ref{tab:time_memory} demonstrates the test time of previous features~\cite{superpoint,r2d2,aslfeat,lbr}, matchers~\cite{superglue, sgmnet}, and our method. Our method (33.2ms) is faster than R2D2 (72.4ms)~\cite{r2d2} and slower than SPP (13.1ms)~\cite{superpoint}, but has higher accuracy. Besides, our method is faster than LBR (9.2+30.1ms)~\cite{lbr}, which uses explicit instances to filter keypoints and advanced matchers including SPP+SPG (13.1+52.2/146.5ms)~\cite{superpoint,superglue} and SPP+SGM (13.1+33.2/97.6ms)~\cite{superpoint,sgmnet}. As the matching method is extensively used for each image pair in mapping and localization processes, SPG/SGMNet are about 18.3/3.3 times slower than NN on Aachen dataset~\cite{visbenchmark,aachen} in the mapping process as discussed in~\cite{sgmnet}. Therefore, our approach could be a good trade off between accuracy and efficiency.

\subsection{Ablation study}
\label{sec:ablation_study}

In Table~\ref{tab:aachen_ablation}, we verify the effectiveness of all components in our network by progressively adding the semantic detection (SD), description (SS), and feature consistency (SF) losses. We also compare results of SS loss with triplet and ranking as intra-class loss. Our baseline is trained with detection scores of SPP~\cite{superpoint} as detector supervision and ap~\cite{aploss} loss for descriptor learning. After adding SD loss, the model performs better especially for night images. The accuracy is further improved by introducing SS loss because it augments the discriminative ability of descriptors with semantics. Compared to triplet loss with carefully tuned margin, ranking loss improves more accuracy as objects' inner diversity can be better retained by optimizing ranks of samples. SF loss enhances the network's ability of embedding semantic information, leading to further improvements. 

\setlength{\tabcolsep}{8.pt}

\begin{table}
	\scriptsize
	\centering
		\begin{tabular}{lcccc}
			\toprule
			SD & SS & SF & Day & Night  \\
			& & & \multicolumn{2}{c}{$(2^\circ,0.25m)/ (5^\circ,0.5m)/ (10^\circ,5m)$} \\	
			\midrule
			\xmark & \xmark & \xmark & 85.4 / 93.6 / 97.9 & 71.2 / 84.3 / 98.4\\
			\cmark & \xmark & \xmark & 87.3 / 94.3 / 97.8 & 72.8 / 88.5 / \textbf{99.5}\\
			\xmark & \cmark (triplet) & \xmark & 87.9 / 95.1 / \textbf{98.9} & 73.8 / 86.9 / 99.0 \\
			\xmark & \cmark (ranking) & \xmark & 87.9 / 95.3 / 98.7 & 74.9 / 89.5 / 99.0 \\
			\cmark & \cmark & \xmark & \textbf{88.2} / 95.5 / 98.8 & 75.9 / 89.0 / \textbf{99.5}\\
			\cmark & \cmark & \cmark & \textbf{88.2} / \textbf{96.0} / 98.7 & \textbf{78.0} / \textbf{92.1} / \textbf{99.5} \\
			\bottomrule
		\end{tabular}
\caption{\textbf{Ablation study.} We test the efficacy of semantic detection (SD), semantic description (SS), and semantic feature consistency (SF) losses. The best results are highlighted.}
\label{tab:aachen_ablation}
\end{table}

\section{Limitations}
\label{sec:limitation}

The first limitation is the hand-defined stability values. A learned stability map from training data could be more robust and further improve the localization accuracy. Besides, semantic labels used in the paper are from ADE20K~\cite{ade20k} and the number of these labels is limited. Fine grained semantic labels~\cite{fgsn} from automatic segmentation might be more reliable in real applications. Moreover, this work focuses mainly on outdoor localization and may not work very well in indoor scenarios due to the significant differences of object classes. Better performance for indoor scenes can be achieved by retraining the model with redefined stability map for indoor objects. 

\section{Conclusions}
\label{sec:conclusion}

In this paper, we implicitly incorporate semantic information into the feature detection and description processes, enabling the model to extract globally reliable features from a single network end-to-end. Specifically, we leverage outputs of an off-the-shelf semantic segmentation network as guidance and adopt a combination of semantic- and feature-aware guidance strategies to enhance the ability of embedding semantic information at training time. Experiments on large-scale visual localization datasets demonstrate that our method outperforms prior local features and gives competitive performance to advanced matchers but has higher efficiency. We argue that our approach could be a good trade-off between accuracy and efficiency. 

\appendix
\section*{\centering Supplementary Material}

In the supplementary material, we first show localization results on Extended CMU-Seasons dataset in Table~\ref{tab:ecmu}. Next, we give more qualitative comparison of previous and our methods on feature extraction and matching in Sec.~\ref{sec_exp_comp}. Then we provide a detailed ablation study of our approach in Sec.~\ref{sec_exp_abla}. Finally, we introduce the process of generating global stability and the architecture of our network in Sec.~\ref{sec:global_stability} and Sec.~\ref{sec:network}, respectively.

\setlength{\tabcolsep}{4.pt}

\begin{table*}
	\tiny
	\centering
		\begin{tabular}{llccccccccccc}
			\toprule
			Group & Method & urban & suburban & park & overcast &sunny &	foliage	& mixed foliage	& no foliage & low sun & cloudy	& snow  \\
			\midrule
			\multirow{3}{*}{C}
			& SIFT~\cite{sift} & 56.9/63.9/70.2	& 37.8/45.3/55.4 & 20.0/24.4/31.7 & 36.1/42.6/50.5 & 30.9/36.3/43.6 & 32.7/38.2/45.7	& 35.5/42.2/51.4 & 59.5/67.5/74.7 & 43.7/50.8/59.2	& 43.0/49.6/58.3 & 46.1/54.2/63.1 \\
			& AS~\cite{as} & 81.0/87.3/92.4	& 62.6/70.9/81.0 & 45.5/51.6/62.0 & 64.1/70.8/78.6 & 55.2/62.3/71.3 & 58.8/65.3/73.9 & 59.2/67.5/77.4 & 83.3/88.9/94.6 & 65.8/73.4/82.8 & 71.6/77.6/84.2 & 73.0/81.0/90.5 \\
			& CSL~\cite{csl} & 71.2/74.6/78.7 & 57.8/61.7/67.5 & 34.5/37.0/42.2 & 52.2/55.4/60.3 & 43.3/46.6/51.9 & 47.0/50.2/55.3	& 52.4/56.1/62.0	& 80.3/83.2/86.6 & 61.7/65.3/70.7 & 63.3/66.3/70.5 & 69.9/73.7/78.7 \\
			
			\midrule
			
			\multirow{2}{*}{S}
			& VLM~\cite{lln} & 17.3/42.5/89.0 &	5.8/19.4/76.1 & 6.6/23.1/73.0 & 11.5/30.8/80.8 & 9.7/27.1/76.1 & 9.5/26.7/77.4 & 10.3/28.4/79.0 & 9.4/30.3/84.6 & 9.3/27.6/79.2 & 9.4/28.0/83.7 & 7.6/27.6/75.9\\
			& SSM~\cite{ssm} & 88.8/93.6/96.3 & 78.0/83.8/89.2 & 63.6/70.3/77.3 & 79.1/84.9/89.7 & 69.2/75.4/81.3 & 73.4/79.1/84.2	& 75.1/81.8/87.9	& 90.9/94.5/97.1 & 78.5/84.5/90.1 & 86.4/90.5/92.9	& 84.1/89.8/94.6\\
			
			\midrule
			
			\multirow{3}{*}{L}
			& SPP~\cite{superpoint} & 89.5/94.2/97.9 & 76.5/82.7/92.7 & 57.4/ 64.4/80.4 &	77.1/82.8/91.8 & 65.1/72.3/86.8 & 69.2/ 75.5/88.3	& 75.2/81.7/90.8 & 88.7/92.8/96.4 & 78.0/83.9/91.8 & 83.4/87.7/94.0 & 80.7/86.6/93.2\\
			
			& D2Net(MS)~\cite{d2-net} & 82.6/94.8/98.4 & 75.9/86.8/93.8	& 66.6/82.6/ 88.6 & 76.3/89.0/94.1 & 68.2/83.8/92.0 & 70.4/85.2/92.5	& 75.8/88.6/93.8 & 86.2/94.4/96.7 & 78.6/89.9/94.4 & 79.1/90.7/95.1 & 82.0/91.1/93.8\\
			
			& R2D2~\cite{r2d2} & 89.7/96.6/98.3	& 76.1/83.8/89.0 & 64.4/72.1/76.5 & 79.9/87.0/90.6 & 70.3/78.3/83.2	& 74.1/81.2/85.6 & 75.7/84.1/87.9 & 86.6/93.3/95.3 & 77.8/85.7/89.3 & 84.1/90.0/92.5	& 79.8/87.6/91.1\\
			
			\midrule
			
			\multirow{3}{*}{M}
			& PixLoc~\cite{patch2pix} & 92.8/95.1/98.5 & 91.9/93.4/95.8	& 84.0 /85.8/90.9	& 90.3/92.2/96.2 & 85.3/88.8/94.0 & 87.1/89.9/94.7 & 90.5/91.9/95.1 & 95.1/95.7/96.8 & 91.2/92.3/94.8 & 93.9/94.8/97.4 & 91.6/92.3/94.0 \\
			
			& AHM~\cite{ahm2019} & 65.7/82.7/91.0 &	66.5/82.6/92.9 & 54.3/71.6/84.1 & 62.8/78.8/89.4 &	56.6/74.5/87.2 & 58.5/75.7/87.8	& 62.9/79.6/89.4	& 72.0/87.7/94.5 & 64.0/81.0/90.2 & 69.4/84.4/92.8 & 61.7/80.6/90.3 \\
			
			& SPP+SPG\cite{superpoint,superglue} & 95.5/98.6/99.3	& 90.9/94.2/97.1 & 85.7/89.0/91.6 & 92.3/95.3/96.9 & 86.1/91.3/94.6 & 88.3/92.5/95.3 & 91.6/94.5/96.2 & 95.4/97.1/98.3 & 91.8/94.4/96.3 & 95.2/97.0/98.0 & 92.3/94.6/96.6 \\
			\midrule
			& Ours & 95.0/97.5/98.6 & 90.5/92.7/95.3 & 86.4/89.1/91.2 & 92.1/94.0/95.8 & 86.3/90.3/93.4	& 87.9/91.0/93.9 & 91.9/94.0/95.5 & 95.3/96.6/97.6 & 92.4/94.4/95.8 & 93.3/94.7/96.3 & 92.9/94.6/96.0\\
			\bottomrule
		\end{tabular}
\caption{\textbf{Localization accuracy on the Extended CMU-Seasons dataset~\cite{visbenchmark}.} Results at error thresholds of $(0.25m, 2^\circ), (0.5m, 5^\circ), (5m, 10^\circ)$ are reported.}
\label{tab:ecmu}
\end{table*}

\section{Analysis of Feature Detection and Matching}
\label{sec_exp_comp}

In this section, we show more qualitative results of keypoints detection and matching in comparison with previous popular local features including SuperPoint~\cite{superpoint}, D2Net~\cite{d2-net}, R2D2~\cite{r2d2}, and ASLFeat~\cite{aslfeat}.

\subsection{Feature detection}
\label{sec:exp_det_comp}

\begin{figure*}[t]
	\centering
	\includegraphics[height=20cm]{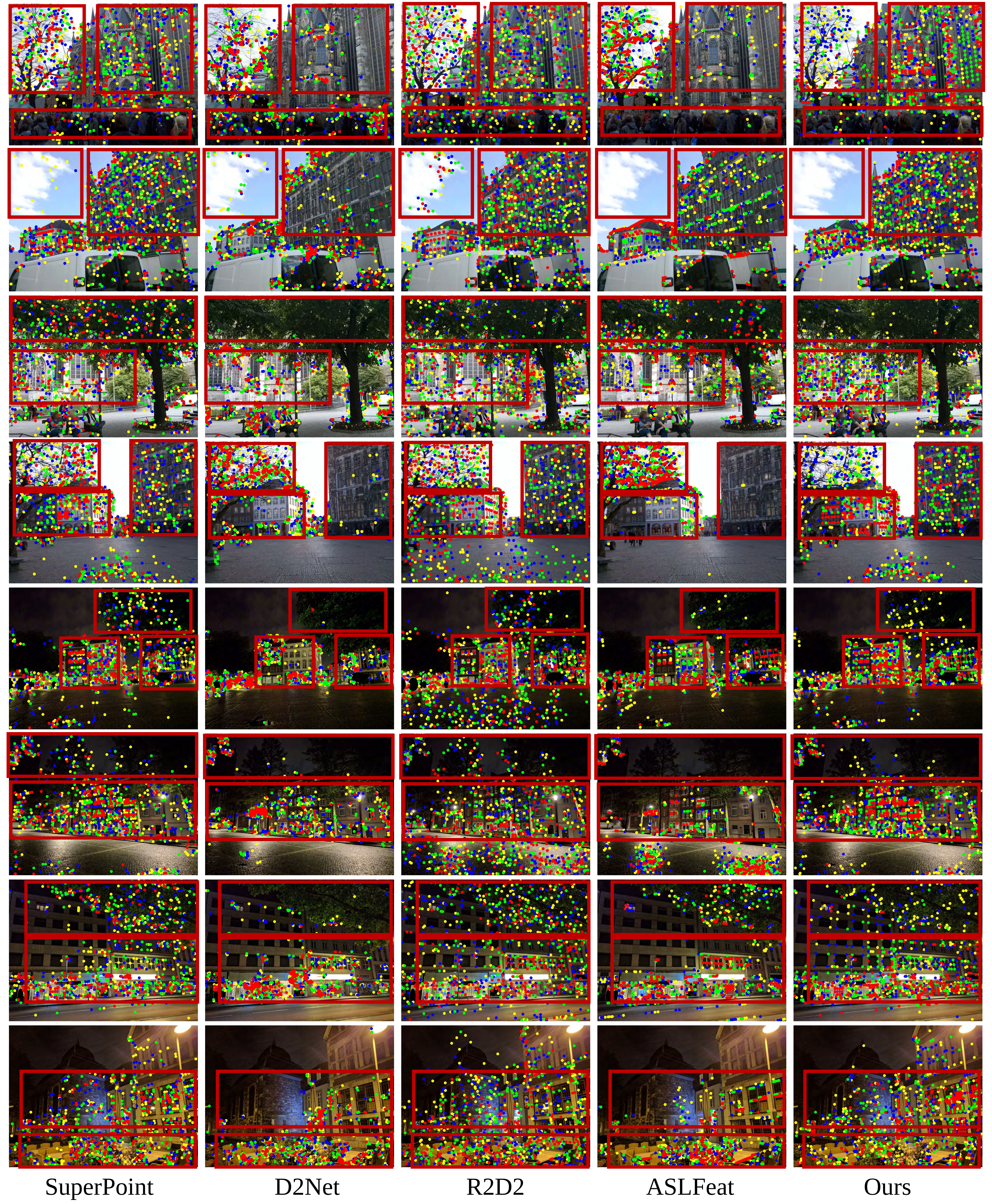}
	\caption{\textbf{Comparison of feature detection.} We show top 1k keypoints with highest scores (high$\rightarrow$low: \colorbox{red}{1-250}, \colorbox{green}{251-500}, \colorbox{blue!70}{501-750}, \colorbox{yellow}{751-1000}) of prior SOTA local features including SuperPoint~\cite{superpoint}, D2Net~\cite{d2-net}, R2D2~\cite{r2d2}, and ASLFeat~\cite{aslfeat}. They are more sensitive to regions with rich textures even those from objects \eg sky, tree, pedestrian, car, which are unstable for long-term localization. By introducing the semantics for reranking keypoints, our model prefers keypoints from stable objects \eg building.}
	\label{fig:det_comp}
\end{figure*}

For each method, we detect top 1k keypoints with highest scores from the query images of Aachen\_v1.1 dataset~\cite{aachen, visuallocalization} at the original resolution and visualize these keypoints with different colors according to their scores (high$\rightarrow$low: \colorbox{red}{1-250}, \colorbox{green}{251-500}, \colorbox{blue!70}{501-750}, \colorbox{yellow}{751-1000}). As shown in Fig.~\ref{fig:det_comp}, we can see that:

\begin{itemize}
	\item D2Net~\cite{d2-net} and ASLFeat~\cite{aslfeat} favor regions with rich textures especially objects such as tress and pedestrians, partially because D2Net and ASLFeat adopt the similar detection strategy: spatial locations with high values of the high-level features. As a result, they detect many keypoints from objects \eg sky, tree, car, pedestrian, which are not useful for long-term localization.
	
	\item R2D2~\cite{r2d2} detects keypoints almost uniformly from the whole image due to its maximization of responses in a fixed sliding window with size of 16$\times$16. Therefore, R2D2~\cite{r2d2} also detects a large number of keypoints from unstable objects.
	
	\item SuperPoint~\cite{superpoint} is a good corner detector. As corners also exist in objects \eg sky, tree, car, pedestrian, SuperPoint~\cite{superpoint} detects many keypoints from the aforementioned unstable objects. 
	
	\item Our detector is partially supersized by results of SuperPoint, so it favors corners as well. Because we rerank the corners with the \textit{stability} of semantic labels, our method prefers to detect keypoints from stable objects \eg building, with more \textcolor{red}{red} keypoints from buildings. Although we can see keypoints from unstable objects, their scores are relatively smaller (with \textcolor{blue!70}{blue} or \textcolor{yellow}{yellow} colors).
	
	\item The distribution of keypoints detected by prior methods and our model indicates that without explicit semantic labels, previous approaches don't perform well of selecting globally reliable keypoints although they are trained to detect keypoints which have strong discriminative ability. 
	
\end{itemize}

\subsection{Feature matching}
\label{sex_exp_matching_comp}

\begin{figure*}[t]
	\centering
	\includegraphics[width=.98\linewidth]{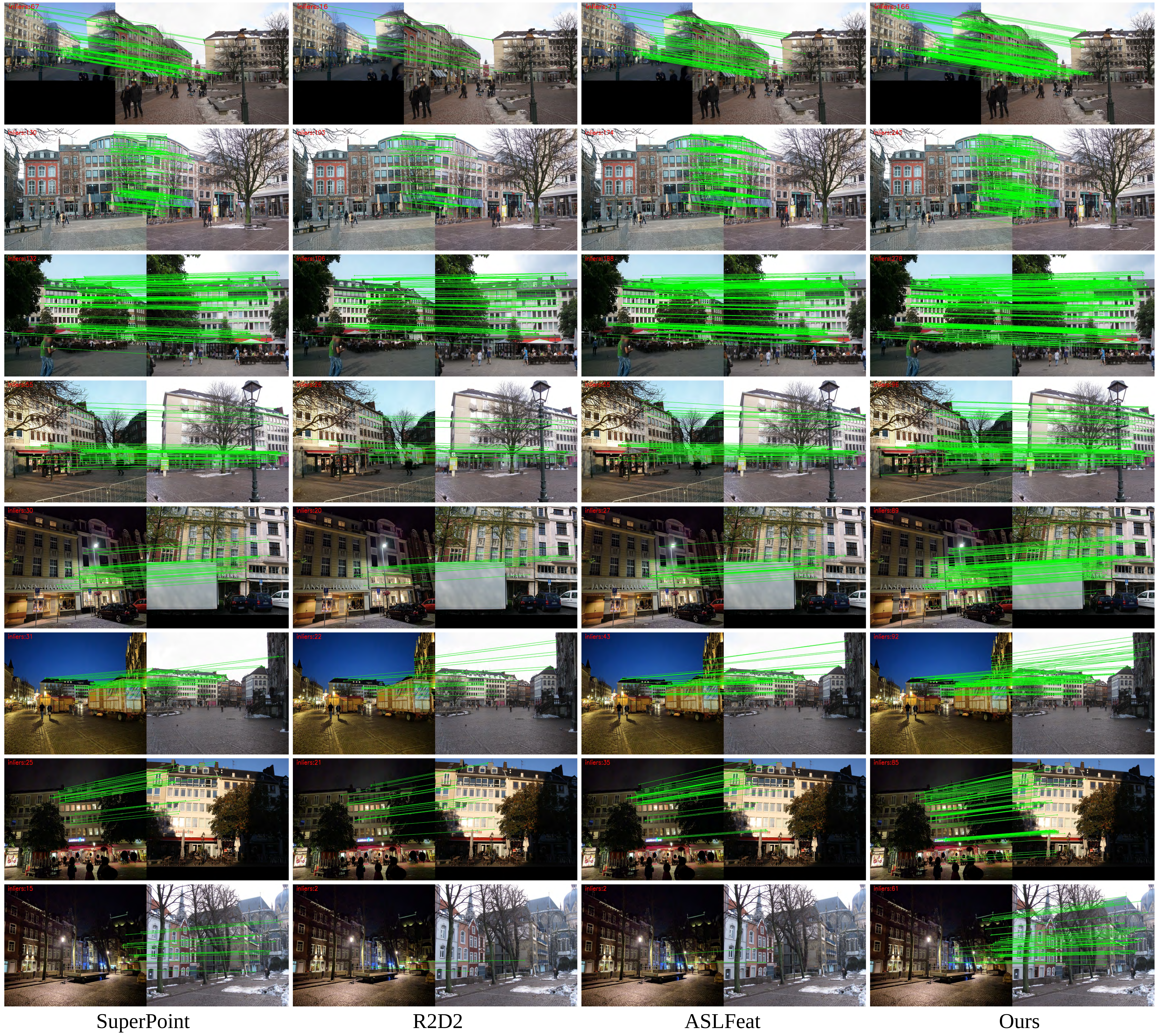}
	\caption{\textbf{Comparison of feature matching.} We show inliers between query and reference images from the Aachen\_v1.1~\cite{aachen,visuallocalization} dataset under challenges of illumination changes, season variations, and dynamic objects. Results of SuperPoint~\cite{superpoint}, R2D2~\cite{r2d2}, ASLFeat~\cite{aslfeat}, and our model are visualized. Compared with prior methods, our model is able to produce more inliers even under extremely challenging conditions when others fail to give enough inliers to guarantee the success of localization.}
	\label{fig:match_comp}
\end{figure*}

We detect 4k keypoints for SuperPoint~\cite{superpoint}, R2D2~\cite{r2d2}, ASLFeat~\cite{aslfeat}, and our method and visualize the inliers between query and reference images with illumination changes, season variations, dynamic objects in the Aachen\_v1.1 dataset~\cite{aachen,visuallocalization}. From Fig.~\ref{fig:match_comp}, we can see that:

\begin{itemize}
	\item For image pairs with small illumination and season changes, almost all methods could give many inliers.
	
	\item For image pairs with season changes or occlusions from trees or dynamic objects \eg car, SuperPoint, R2D2, and ASLFeat give fewer inliers than our model.
	
	\item For extremely challenging image pairs with illumination changes, season variations, and high occlusions of trees, almost all prior approaches fail to give enough inliers, resulting in the failure of localization. However, our method is still able to find enough inliers from robust regions. We analyze the reasons of improvements in Sec.~\ref{sec_exp_abla}.
\end{itemize}

\section{Ablation Study of Feature Detection and Matching}
\label{sec_exp_abla}

In this section, we verify the efficacy of the proposed semantic-aware detection (SD), semantic-aware description (SS), and semantic-consistency (SF) losses by visualizing the detection and matching results. The base model is trained with results of SuperPoint~\cite{superpoint} as supervision and a general ap loss~\cite{aploss} for descriptor learning as R2D2~\cite{r2d2}. Our full model comprises SD, SS, and SF three components.

\subsection{Ablation study of detection}
\label{sec_exp_det_abla}
As in Sec.~\ref{sec:exp_det_comp}, we visualize 1k keypoints with the highest scores and show them with different colors according to their scores (high$\rightarrow$low: \colorbox{red}{1-250}, \colorbox{green}{251-500}, \colorbox{blue!70}{501-750}, \colorbox{yellow}{751-1000}). As shown in Fig.~\ref{fig:det_comp}, we can see the effectiveness of SD, SS, and SF in detail:

\begin{itemize}
	\item Our base model performs closely to SuperPoint~\cite{superpoint} (as shown in Fig.~\ref{fig:det_comp}) with high response to corners as the detector is partially supervised with results of SuperPoint~\cite{superpoint}. Meanwhile, the base model is also sensitive to unstable objects \eg sky, tree, pedestrian, and car.
	
	\item The SD loss (W/ SD) is the key to rerank the keypopints. With SD loss, keypoints from unstable objects \eg sky, car, pedestrian are suppressed. Keypoints from trees have lower score (with color of \textcolor{blue}{blue} or \textcolor{yellow}{yellow}) and keypoints from stable objects \eg building are favored (with color of \textcolor{red}{red}).
	
	\item The SS loss doesn't contribute to the detection process, so it shows the similar results as the base model, which again indicates that the importance of explicit semantic labels to detection as discussed in Sec.~\ref{sec:exp_det_comp},
	
	\item The full model with SF incorporated performs better than the model W/ SD, as it further enhances the ability of our model in learning semantic-aware features.
	
\end{itemize}

\subsection{Ablation study of feature matching}
\label{sec_exp_matching_abla}

\begin{figure*}[t]
	\centering
	\includegraphics[width=1.\linewidth]{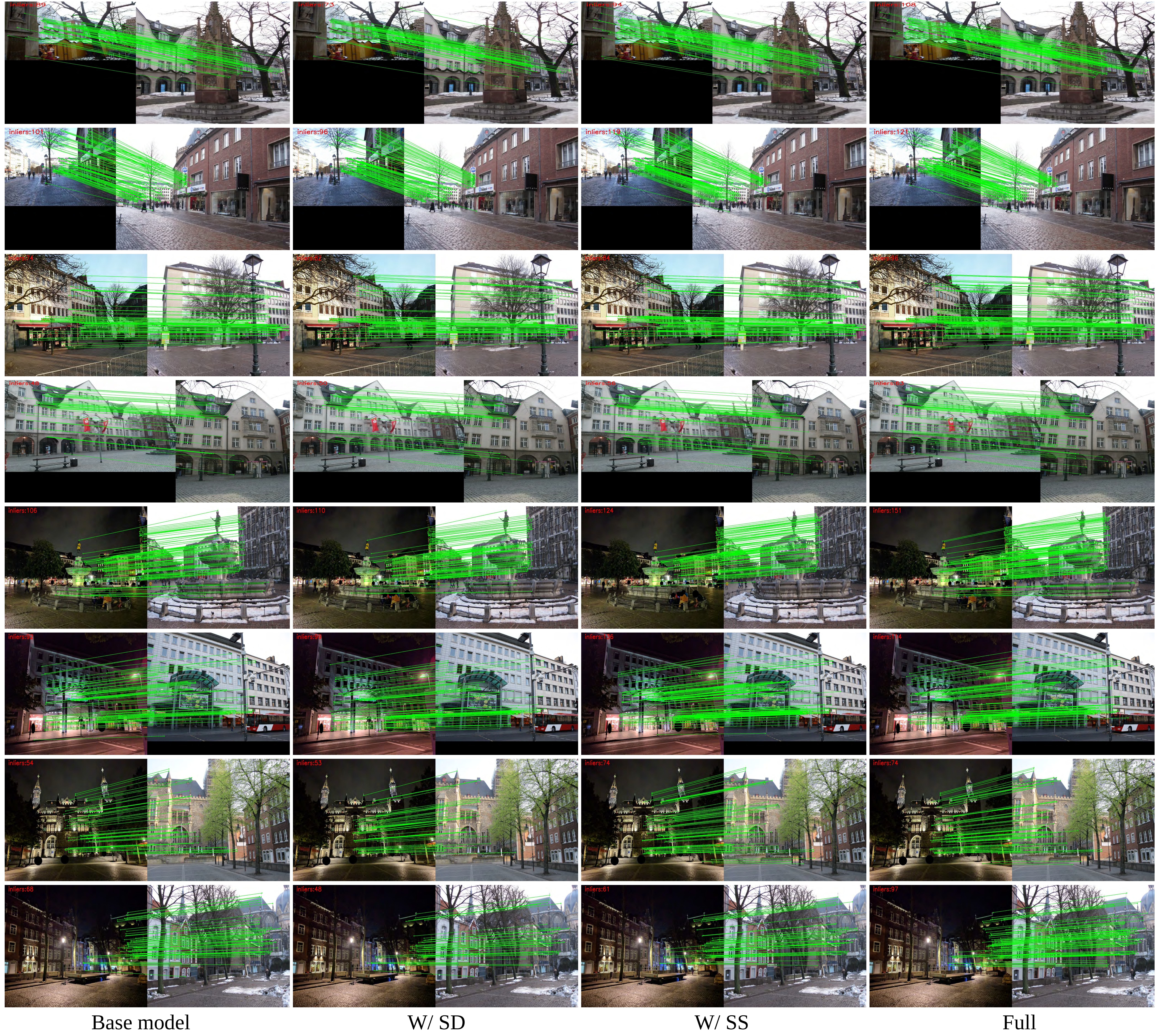}
	\caption{\textbf{Ablation study of feature matching.} We show the inliers between query and reference images from the Aachen\_v1.1~\cite{aachen,visuallocalization} dataset under challenges of illumination changes, season variations and dynamic objects. Results of the base model, with SD loss (W/ SD), with SS loss (W/ SS) and the full model (with SD, SS, SF) are visualized. SD loss slightly improves the matching as it focus mainly the detection process. SS loss effectively augments the matching accuracy by introducing semantic labels. Results of SS loss are further improved by the full model, which has an additional SF loss to enhance the model's ability of learning semantic-aware features.}
	\label{fig:match_abla}
\end{figure*}

We additionally visualize the effectiveness of SD, SS, SF losses in feature matching. From Fig.~\ref{fig:match_abla}, we can see that:

\begin{itemize}
	\item Benefiting from the corner detector and ap loss, the base model is already able to give promising performance in comparison with previous methods~\cite{superpoint, r2d2, aslfeat}.
	
	\item The SD loss (W/ SD) marginally improves the matches possibly because those reranked keypoints from stable objects don't have strong discriminative ability by purely adopting ap loss over all keypoints.
	
	\item The SS loss (W/ SS) effectively solves the limitation of SD loss, as it augments the discriminative ability of descriptors with semantics.
	
	\item The full model gives the best performance because it combines the advantages of SD, SS, and SF losses.
\end{itemize}

\setlength{\tabcolsep}{2.pt}

\begin{table}[t]
		\footnotesize
	\centering
	\begin{threeparttable}
		\begin{tabularx}{\linewidth}{l|c}
			\toprule
			Category & Semantics \\
			\midrule
			\textbf{Volatile} & \parbox[c]{6.5cm}{sky, mountain, curtain, water, sea, mirror, rug, field, bathtub, stand, sand, sink, river, hill, bench,light,dirt,land,fountain, swimming pool, waterfall, lake} \\
			\midrule
			
			\textbf{Dynamic} & \parbox[c]{6.5cm}{person, automobile, boat, truck} \\
			\midrule
			
			\textbf{Short-term} & \parbox[c]{6.5cm}{tree, grass, plant, flower, palm, airplane, van, ship, minibike, bike, shower} \\
			\midrule
			
		 	\textbf{Long-term} & \parbox[c]{6.5cm}{wall, building, floor, ceiling, road, bed, window, cabinet, sidewalk, ground, door, chair, painting, sofa, shelf, house, armchair, seat, fence, rock, wardrobe, lamp, rail, cushion, box, pillar, signboard, chest, counter, skyscraper, fireplace, grandstand, path, stairs, runway, case, table, pillow, screen, stairway, bridge, bookcase, toilet, book, countertop, stove,  kitchen, computer, swivel, bar, arcade, hovel, tower, chandelier, sunshade, streetlight, booth, television, clothes, pole, bannister,  escalator, ottoman, bottle, buffet, poster, stage, conveyer,  canopy, washer, toy, stool, cask, basket, tent, bag, cradle, oven, ball, food, step, tank, trade name, pot, dishwasher, screen, blanket, sculpture, hood, sconce, vae, traffic light, tray, dustbin, plate, monitor, bulletin, glass, clock, flag} \\
			\bottomrule
		\end{tabularx}
	\end{threeparttable}
\caption{\textbf{Stability map of different labels}. All semantic labels are categorized into four groups denoted as \textit{Volatile}, \textit{Dynamic}, \textit{Short-term}, and \textit{Long-term} according to their reliability in the visual localization task.}
\label{tab:gstability}
\end{table}

\begin{figure}[b]
	\centering
	\includegraphics[width=1.\linewidth]{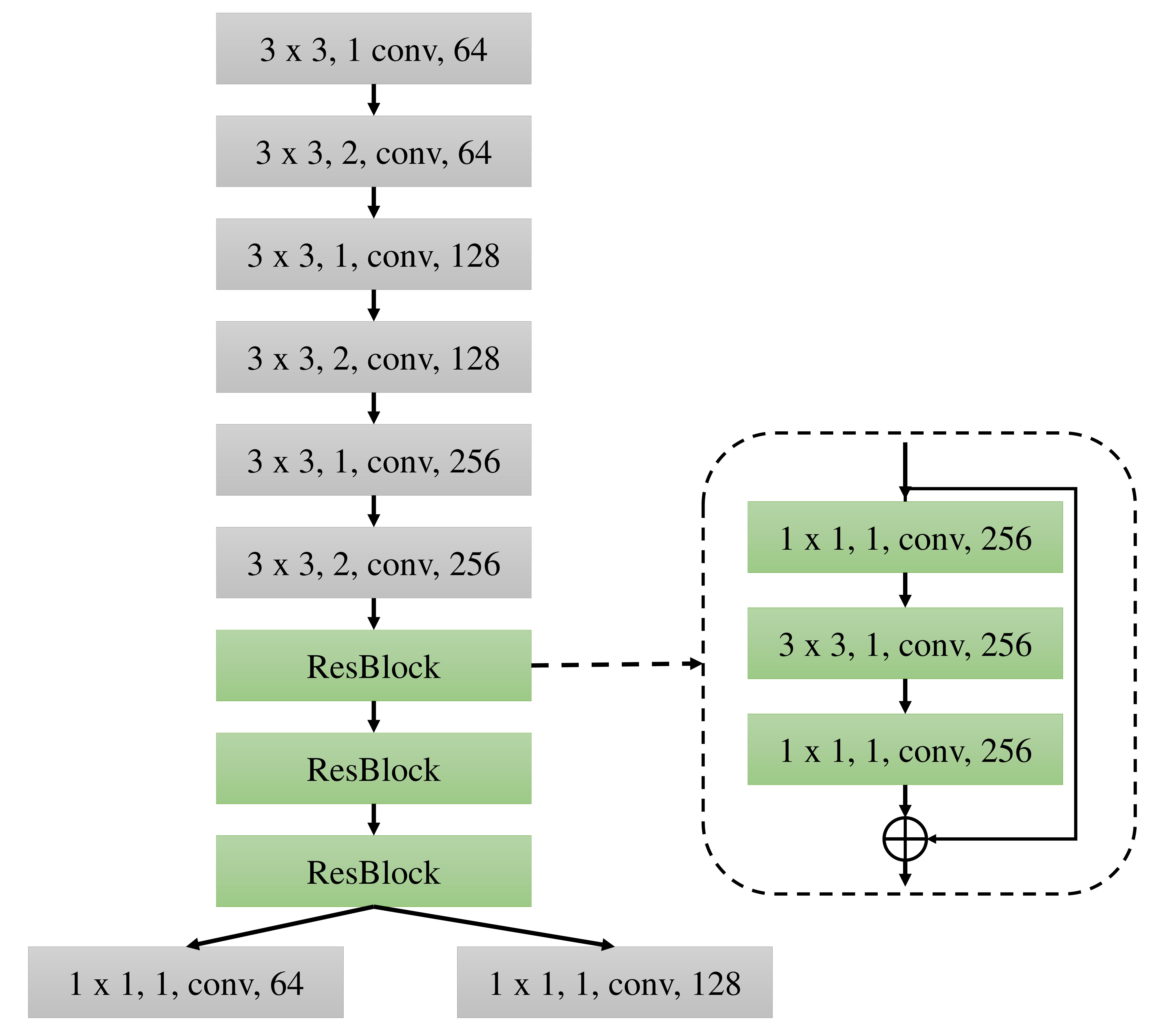}
	\caption{\textbf{Architecture of the network.} We adopt 6 Convolution layers with kernel size of $3\times3$ to generate high-level features with $8 \times$ downsampling (implementation by using stride of 2). Then 3 ResBlocks~\cite{resnet} are followed to further enhance the ability of the model.}
	\label{fig:architecture}
\end{figure}

\begin{figure*}[t]
	\centering
	\includegraphics[height=20cm]{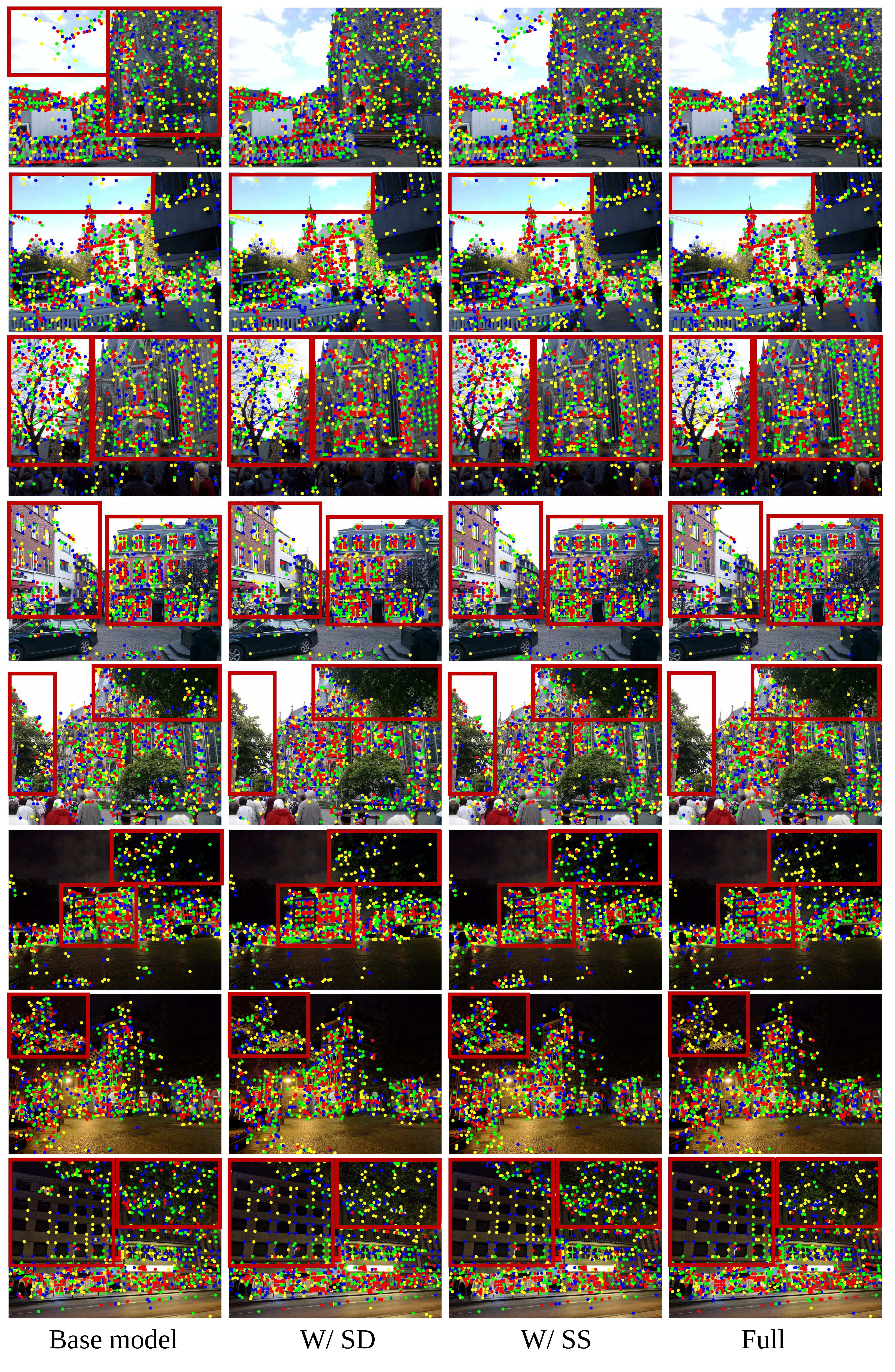}
	\caption{\textbf{Ablation study of feature detection.} We show top 1k keypoints with the highest scores (high$\rightarrow$low: \colorbox{red}{1-250}, \colorbox{green}{251-500}, \colorbox{blue!70}{501-750}, \colorbox{yellow}{751-1000}) of our base model, model with SD loss (W/ SD), SS loss (W/ SS) and the full model (with SD, SS, SF). The base model is more sensitive to regions with rich corners as SuperPoint~\cite{superpoint}. SD loss effectively mitigates this problem by introducing semantic-aware detection loss. SS loss focus mainly on descriptor learning, so it gives similar results to the base model. The full model additionally introduces SF loss, which further enhances the detection process.}
	\label{fig:det_abla}
\end{figure*}

\section{Global stability map generation}
\label{sec:global_stability}
During the training process, we utilize UperNet~\cite{upernet} with ConvNet~\cite{convnet} as encoder trained on ADE20k~\cite{ade20k} dataset to provide semantic segmentation labels and high-level features for semantic-wise and feature-wise guidance, respectively. There are 150 labels in total which are categorized into 4 groups as shown in Table~\ref{tab:gstability}. Since large-scale localization happens mainly in outdoor environments, only several objects such as sky, water, pedestrian, car, tree, plant, and building are frequently used.

\section{Network}
\label{sec:network}
Alike to SuperPoint~\cite{superpoint}, we adopt 8 times downsampling to reduce the resolution of high-dimension features, making the model efficient at test time. To increase the representation ability of our model, we introduce 3 ResBlocks~\cite{resnet}. Details of the network are shown in Fig.~\ref{fig:architecture}.


{\small
\bibliographystyle{ieee_fullname}
\bibliography{egbib}
}

\end{document}